\pdfoutput=1

\documentclass[11pt]{article}
\usepackage[dvipsnames]{xcolor}
\definecolor{skyblue}{RGB}{135,206,235}

\usepackage{color}

\usepackage[final]{acl}

\usepackage{times}
\usepackage{latexsym}

\usepackage[T1]{fontenc}

\usepackage[utf8]{inputenc}

\usepackage{microtype}

\usepackage{inconsolata}

\usepackage{graphicx}
\usepackage{subfig}
\usepackage{times}
\usepackage{soul}
\usepackage{url}
\usepackage{amssymb,amsfonts}
\usepackage[utf8]{inputenc}
\usepackage{amsmath}
\usepackage{amsthm}
\usepackage{booktabs}
\usepackage{algorithm}
\usepackage{algorithmic}
\usepackage{adjustbox} 
\usepackage[switch]{lineno}
\usepackage{caption}
\usepackage{colortbl}

\title{VistaWise\includegraphics[scale=0.12]{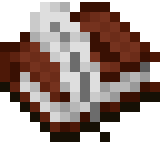}: Building Cost-Effective Agent \\ with Cross-Modal Knowledge Graph for Minecraft}

\author{Honghao Fu$^{1, 2\ddag}$\thanks{The work was done during an internship at HKUST(GZ).}, Junlong Ren$^{1\ddag}$, Qi Chai$^{1}$, 
        \textbf{Deheng Ye$^{3}$, Yujun Cai$^{2}$, Hao Wang$^{1}$\thanks{Corresponding Author. $^{\ddag}$Equal Contribution.}} \\
        $^1$The Hong Kong University of Science and Technology (Guangzhou)\\
        $^2$University of Queensland, 
        $^3$Tencent \\
        \texttt{honghao.fu@uq.edu.au}, \texttt{\{jren686, qchai315\}@connect.hkust-gz.edu.cn}\\ 
        \texttt{dericye@tencent.com}, \texttt{yujun.cai@uq.edu.au}, \texttt{haowang@hkust-gz.edu.cn}}

\begin{document}
\maketitle
\begin{abstract}
Large language models (LLMs) have shown significant promise in embodied decision-making tasks within virtual open-world environments. Nonetheless, their performance is hindered by the absence of domain-specific knowledge. Methods that finetune on large-scale domain-specific data entail prohibitive development costs. This paper introduces VistaWise, a cost-effective agent framework that integrates cross-modal domain knowledge and finetunes a dedicated object detection model for visual analysis. It reduces the requirement for domain-specific training data from millions of samples to a few hundred. VistaWise integrates visual information and textual dependencies into a cross-modal knowledge graph (KG), enabling a comprehensive and accurate understanding of multimodal environments. We also equip the agent with a retrieval-based pooling strategy to extract task-related information from the KG, and a desktop-level skill library to support direct operation of the Minecraft desktop client via mouse and keyboard inputs. Experimental results demonstrate that VistaWise achieves state-of-the-art performance across various open-world tasks, highlighting its effectiveness in reducing development costs while enhancing agent performance.
\end{abstract}

\section{Introduction}

The development of agents in open-world environments is widely regarded as a promising avenue for advancing artificial general intelligence~\cite{Odyssey,generalist}. These agents have to navigate intricate environments and make decisions under conditions of uncertainty. Among the various platforms for developing such agents, Minecraft has emerged as a prominent virtual environment, owing to its open-ended design and the extensive range of task possibilities~\cite{Voyager}. 
Early efforts~\cite{VPT} within Minecraft focused on training visual models using reinforcement learning in simulators~\cite{MineRL,MineDOJO}.
However, these approaches face challenges related to systematic exploration~\cite{enhancedP}, interpretability~\cite{proto}, and generalization~\cite{Voyager}.

With the advancement of Large Language Models (LLMs), applying LLM agents on Minecraft has 
become a prominent trend~\cite{GITM,DEPS}.
However, due to limitations in environmental perception and grounding, LLM-based agents in Minecraft often rely on environmental APIs (e.g., MineFlayer\footnote{https://github.com/PrismarineJS/mineflayer}) to obtain accurate textual descriptions of the environment and execute actions~\cite{Voyager,STEVE}. This reliance may hinder generalization, as not all virtual environments provide such APIs. Additionally, directly utilizing APIs for high-level actions (e.g., chopping logs) restricts the autonomy and potential of LLM-based agents~\cite{GROOT}. In contrast, an ideal agent should rely solely on visual cues for reasoning and perform tasks using low-level and hybrid action spaces, which involve mouse and keyboard (MNK) operations, more closely mimicking human behavior.

Building on this objective, several studies~\cite{JARVIS-1,ROCKET-1,OmniJARVIS} propose the use of Multimodal LLMs (MLLMs)~\cite{liu2025step1x,zhang2025tokenswap} for reasoning based on visual information, subsequently enhancing visual policies with the reasoning outcomes. In parallel, to mitigate performance degradation or hallucinations~\cite{mei2024not,wang2025text,wang2025cure,li2025texture} resulting from the LLM's lack of domain-specific knowledge in virtual environments, finetuning the LLM with knowledge tailored to the target virtual world has become a widely adopted strategy~\cite{DEPS,OmniJARVIS,JARVIS-1,STEVE,MP5}. However, several challenges exist: (1) The reliance on additional visual policies to predict actions diminishes the LLM's decision-making potential. Moreover, this approach usually models visual~\cite{gu2025acl,gu2025sfir} input globally, incorporating irrelevant noise that results in unstable predictions for MNK operations~\cite{ROCKET-1}. 
(2) General visual policies or LLMs may underperform due to the lack of domain-specific knowledge. However, finetuning them with large-scale domain-specific data needs prohibitive development costs.

\begin{figure}[t]
  \centering
  \setlength{\abovecaptionskip}{5pt} 
  \includegraphics[width=\linewidth]{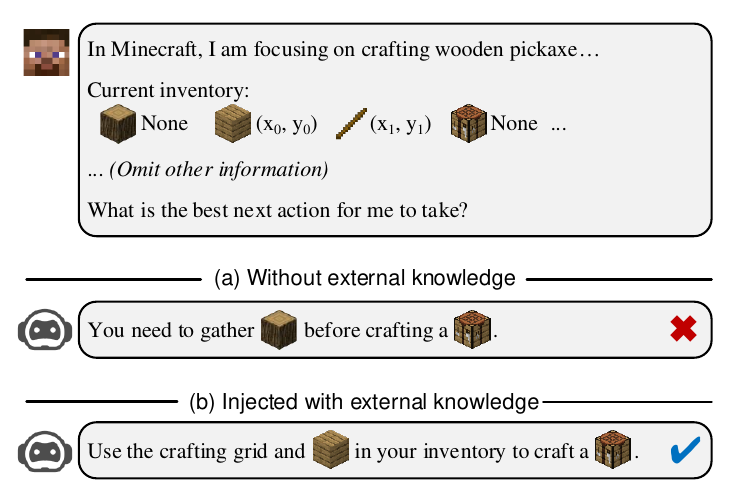}
  \caption{\textbf{The motivation of injecting external knowledge.} (a) LLMs may generate incorrect dependencies due to a lack of domain-specific knowledge in the virtual world; (b) Injecting external knowledge enables LLMs to generate more accurate responses.}
  \label{KG-inj}
    \vspace{-0.15in}
\end{figure}

To address these problems, we propose VistaWise, a cost-effective agent with a cross-modal knowledge graph for Minecraft.
VistaWise incorporates an object detection model to identify visual entities, effectively minimizing irrelevant information and noise. This model is the sole component in VistaWise that requires finetuning on domain-specific data. It is efficiently finetuned with fewer than 500 annotated frames extracted from gameplay videos, thereby significantly reducing the training cost for embedding domain knowledge into the visual modality. 

VistaWise utilizes an LLM as the policy to directly predict actions, harnessing the LLM’s reasoning and decision-making capabilities. 
To mitigate hallucinations (e.g., as shown in Figure~\ref{KG-inj}), we aim to integrate domain-specific knowledge into the textual modality through external knowledge retrieval~\cite{RAG}, thus eliminating the need for additional finetuning of the LLMs.
Inspired by the efficacy of knowledge graphs (KGs) in reducing hallucinations~\cite{GraphRAG}, we construct a textual KG using online textual information as the external knowledge base to provide factual dependencies. 
Given the multimodal nature of tasks in Minecraft, the visual information extracted by the object detection model is further embedded into the graph, enabling the construction of the cross-modal knowledge graph that enhances the agent’s comprehensive and accurate understanding of the environment. To mitigate the inference cost of LLMs due to information redundancy, we further design a retrieval-based graph-pooling strategy for efficiently accessing the cross-modal information stored within the graph structure.

To enhance the agent's dynamic decision-making capabilities, we integrate a concise task description, Chain-of-Thought (CoT) reasoning~\cite{CoT}, and a memory module. To facilitate action execution, we develop a skill library that references human players' MNK behaviors using PyAutoGUI, from which the LLM autonomously generates parameters based on visual cues. This approach empowers the agent to directly control the game on the desktop, eliminating the need for simulation environments or environmental APIs. 
Our key contributions are summarized as follows:
\begin{itemize}
\setlength{\itemsep}{0pt}
\setlength{\parsep}{0pt}
\setlength{\parskip}{0pt}
\item We propose a cost-effective agent framework that incorporates multimodal domain-specific knowledge by finetuning an object detection model and externally retrieving textual knowledge, reducing training data requirements from millions to only a few hundred samples.
\item We integrate visual and textual information via a KG to construct cross-modal representations, enhancing the agent's understanding of multimodal tasks. We also design a pooling strategy termed retrieval-based pooling to extract information from the introduced KG.
\item By directly leveraging the LLM's reasoning and decision-making capabilities, VistaWise outperforms other non-API-based baselines in complex tasks, achieving a success rate of 33\% in obtaining diamonds, surpassing the previous state-of-the-art rate of 25\%.
\end{itemize}

\section{Related Work}
\subsection{Agents in Minecraft}

\noindent\textbf{Agents based on vision models.}
MineRL~\cite{MineRL} and MineDOJO~\cite{MineDOJO} offer simulation platforms for developing virtual agents in Minecraft. Early agents utilize visual models trained through reinforcement or imitation learning as policies. VPT~\cite{VPT} employs internet-scale pretraining for sequential decision-making via semi-supervised imitation learning. STEVE-1~\cite{STEVE-1} extends VPT by incorporating additional prior knowledge~\cite{fu2024dp} derived from task descriptions and task execution videos. However, these methods face challenges related to systematic exploration, interpretability, and generalization~\cite{Voyager}.

\noindent\textbf{Agents based on LLMs.}
Compared to traditional visual models, LLMs excel in complex cognitive tasks and causal reasoning~\cite{mei2025survey,mei2024slang,ren2025wamo,ren2025diversified,ren2025exploiting,ti2025towards,zhang2025improving,li2024drs,li2024vulnerability,10888525,zhang2025multimindenhancingwerewolfagents}, driving the development of Minecraft agents based on LLM policies~\cite{GITM,DEPS}. To address limitations of LLMs in visual~\cite{gu2023orsi,gu2025optical,meng2025wavelet} capabilities, Voyager~\cite{Voyager} proposes to leverage environmental APIs to obtain precise data and execute high-level actions. With the emergence of MLLMs~\cite{fu2025brainvis,zhang2024defending,zhang2025tuning,zhang2026test}, subsequent works such as STEVE~\cite{STEVE}, LARM~\cite{LARM}, MP5~\cite{MP5}, LLaMA-Rider~\cite{LLaMA-Rider}, and Odyssey~\cite{Odyssey} further enhance agent performance by incorporating visual inputs and finetuning MLLMs with domain-specific data. However, these methods rely on certain APIs to gather information and execute actions, facing challenges in generalization, as not all environments provide such APIs.

\noindent\textbf{Agents integrating LLMs and vision models.}
To enable LLM-based agents to execute actions solely based on visual input, without reliance on APIs, JARVIS-1~\cite{JARVIS-1} introduces the use of MLLMs for long-term planning. ROCKET-1~\cite{ROCKET-1} further incorporates priors from MLLMs and SAM \cite{SAM} to train a visual policy. 
However, these approaches incur substantial costs in data collection and training. In this paper, we offer a more efficient framework by minimizing the complexity of these processes.

\subsection{LLMs with Knowledge Graph}
LLMs have demonstrated remarkable language understanding and zero-shot transfer abilities across various natural language processing tasks. However, they still lack up-to-date or domain-specific knowledge \cite{wang2023selfconsistency}, which hinders their generalization.
Recent works have proposed to adopt the knowledge graph (KG)~\cite{wang2021mixup,fu2023sgcn,xiong2025mapping} to provide up-to-date and structured domain-specific knowledge to improve LLM reasoning on knowledge-intensive tasks \cite{pan2024unifying}.
Many methods \cite{zhang2022subgraph,jiang2023unikgqa,mei2024hiddenguard,baek2023knowledge,gu2025hdtcnet,jiang2023reasoninglm} combine LLMs with KGs through converting retrieved relevant knowledge from KGs to textual prompts for LLMs. Nevertheless, the effectiveness of these methods mainly depends on the quality of the retrieved knowledge.
To retrieve more accurate and reliable knowledge from KGs, recent works \cite{jiang2023structgpt, sun2024thinkongraph, jiang2024kg, wang2025reasoning} utilize LLMs to enhance reasoning on KGs.

\begin{figure*}[t]
  \centering
    \setlength{\abovecaptionskip}{5pt}
  \includegraphics[width=\linewidth]{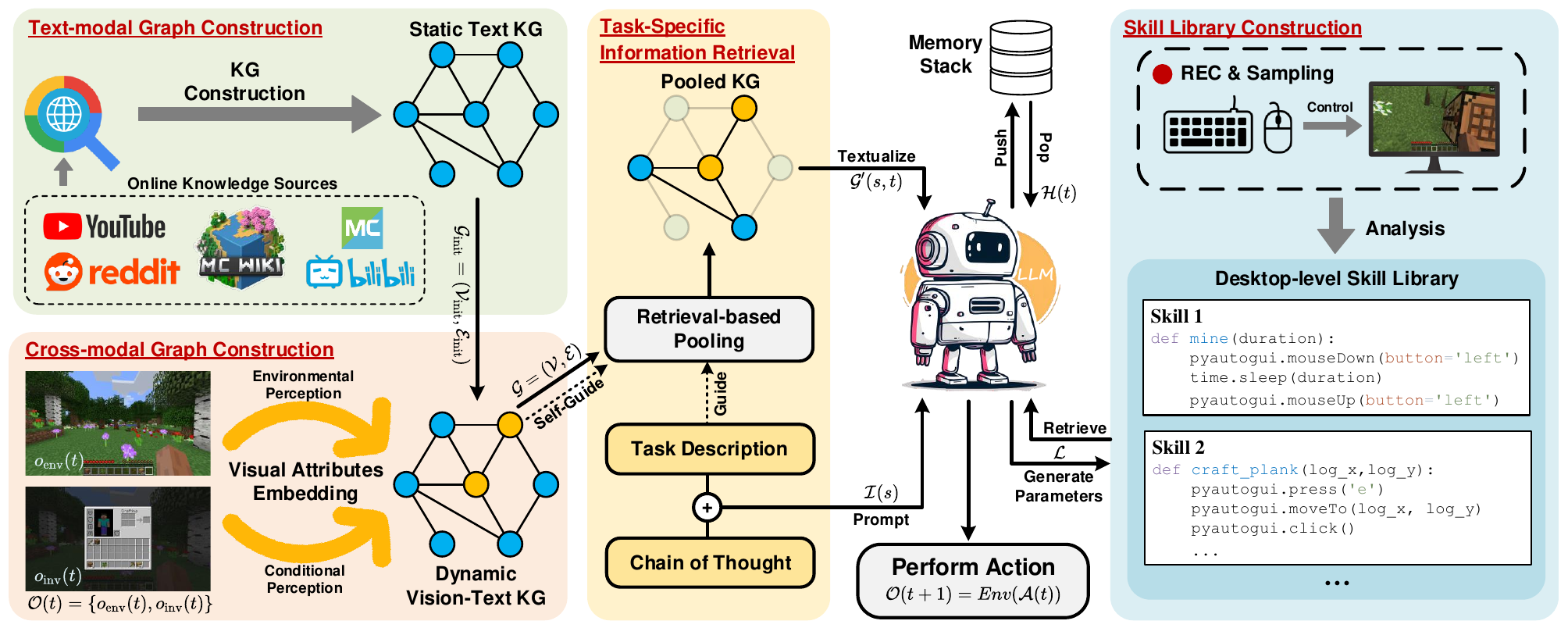}
  \caption{\textbf{The overview of VistaWise.} VistaWise is based on an LLM and incorporates three graph-based processes: (1) \textbf{text-modal graph construction}, integrating external textual domain knowledge via a lightweight KG to establish factual dependencies and mitigate hallucinations; (2) \textbf{cross-modal graph construction}, embedding real-time visual information from a dedicated object detection model into the text-modal graph, forming a vision-text graph with dynamic visual attributes; (3) \textbf{task-specific information retrieval}, utilizing a retrieval-based pooling strategy to extract task-related information from the vision-text graph, guided by both the task-specific prompt and the real-time visual attributes of the graph. Furthermore, VistaWise comprises two interaction modules: (i) a \textbf{desktop-level skill library}, allowing the agent to act in the Minecraft desktop client via MNK operations, with action parameters generated autonomously by the LLM; (ii) a \textbf{memory stack}, storing and querying decision history to support reasoning. At each timestep, the agent makes decisions and executes actions based on the prompt, retrieved information, memory, and skill library, altering the game environment to advance the task.}
  \label{pipe}
  \vspace{-0.1in}
\end{figure*}

\section{Proposed Method}
\label{method}

\subsection{Overview}
Figure~\ref{pipe} shows the framework of VistaWise, which leverages visual inputs and desktop-level control to interact with Minecraft. VistaWise is built around an LLM and consists of three graph-based processes: (1) text-modal graph construction, (2) cross-modal graph construction, (3) task-specific information retrieval, and two interaction modules: (i) a desktop-level skill library and (ii) a memory stack.

\nocite{liu2025correlationcausationmaxpoolingbasedmultiinstance,liu2025hacsurv}

\subsection{Cross-modal Information Integration}
\label{VPPS}
\noindent\textbf{Text-modal graph construction.} As shown in Figure~\ref{KG-inj}(a), the agent may misinterpret entity dependencies and suffer from hallucinations due to a lack of domain-specific knowledge, resulting in incorrect reasoning during crafting tasks. While finetuning LLMs to incorporate new knowledge is a viable approach, it suffers from significant costs related to data collection and additional training. An alternative and widely recognized solution is to provide factual support via an external knowledge base~\cite{RAG,RAG2}, where a structured KG is an effective way to provide factual dependencies~\cite{KGRAG,GraphRAG,GRAG,GraphRAG2}. We structure online textual knowledge into a KG $\mathcal{G}_\text{init} = (\mathcal{V}_\text{init}, \mathcal{E}_\text{init})$, where entity nodes $\mathcal{V}_\text{init}$ (e.g., ``Player,'' ``Tools,'' ``Iron Ingot'') and their relationships $\mathcal{E}$ (e.g., ``includes,'' ``can be used to mine,'' ``is used to craft'') capture factual dependencies. In practice, we find that entity names alone are sufficient for the LLM to understand these dependencies. Therefore, we remove other information (e.g., background knowledge) of the nodes, creating a more lightweight representation. It reduces the external knowledge injected, thus decreasing the computational costs while maintaining comparable performance as indicated in Table~\ref{AEA} and Sec. \ref{sec::Inference Costs}.

\noindent\textbf{Visual perception.} To predict the next action, the agent must not only comprehend the factual dependencies within the virtual world but also perceive entities in the observable space. 
In order to accurately perceive the environment and mitigate token overhead from visual inputs, we utilize a dedicated object detection model to replace the object grounding and detection functions of MLLMs, extracting real-time entity information (e.g., coordinates, bounding boxes) from the observable space.
We classify the entities into two categories: environmental entities, which represent the spatial distribution of the environment, and conditional entities, which can be used to track the progress of the agent's task. We define the observable space for the agent at timestep $t$ as $\mathcal{O}(t)=\{o_\text{env}(t), o_\text{inv}(t)\}$, where $o_\text{env}$ and $o_\text{inv}$ represent the environment and inventory spaces in Minecraft. The goal is to enable the agent to accurately perceive the spatial locations of interactive or semantically significant visual entities within $\mathcal{O}$. The agent's perception is based on two parallel processes: one for environmental entities and another for conditional entities, both supported by an object detection model $D$. To model the virtual environment, we define the environmental entities in $o_\text{env}(t)$ as $E(t) = \{e_\text{t,1}, e_\text{t,2}, \dots, e_\text{t,n}\}$ (e.g., trunks, lava, ores). The agent interacts with these entities to gather resources and modify the environment to achieve the goal task. For any observable entity $e$ at any timestep, the agent obtains its desktop-view information $D(o_\text{env}, e) = \{x_e, y_e, w_e, h_e \}$, where $(x_e, y_e)$ represents the center coordinates of $e$, and $w_e$ and $h_e$ denote the width and height of its bounding box. Let $V_\text{env}$ represent all the information related to $E$ in $o_\text{env}$ accessible through $D$. The $V_\text{env}$ at timestep $t$ is expressed as:
\vspace{-5pt} 
\begin{equation}
\resizebox{1\linewidth}{!}{
$
V_\text{env}(t) = \{ D(o_\text{env}(t), e_\text{t,i}) \mid e_\text{t,i} \in E(t), i = 1, 2, \dots, n \}.
$
}
\end{equation}
Moreover, we define the visual entities in $o_\text{inv}(t)$ (e.g. item icons) as conditional entities $C(t) = \{c_\text{t,1}, c_\text{t,2}, \dots, c_\text{t,m}\}$. For any $c$ at any timestep, the information the agent can observe is $D(o_\text{inv}, c) = \{x_c, y_c\}$, where $(x_c, y_c)$ are the center coordinates of $c$. Agent tracks task progress and assesses whether conditions for specific actions (e.g., crafting) are met based on $C$. Let $V_\text{inv}$ denote all information related to $C$ in $o_\text{inv}$ accessible through $D$. The $V_\text{inv}$ at timestep $t$ is expressed as:
\vspace{-5pt} 
\begin{equation}
\resizebox{1\linewidth}{!}{
$
V_\text{inv}(t) = \{ D(o_\text{inv}(t), c_\text{t,i}) \mid c_\text{t,i} \in C(t), i = 1, 2, \dots, m \}.
$
}
\end{equation}
\noindent In addition, a significant challenge is assessing the distance between the agent and $e$ based on visual information. It is primarily raised by the distribution gap between real-world and virtual environment visuals, which renders deep estimation models trained on real-world data ineffective in capturing the spatial distribution of $o_\text{env}$. Drawing inspiration from how human players empirically estimate distances based on the size of visual entities, we use $w_e$ and $h_e$ to determine whether the agent is within the interaction range of $e$, employing empirical thresholds $k_w$ and $k_h$, respectively. This enables the agent to perceive spatial distance in a coarse-grained manner.

\noindent\textbf{Cross-modal graph construction.} 
The real-time visual information is embedded as attributes of the corresponding entity nodes in the static KG that describes factual relationships. This visual attributes embedding process updates the text KG with static dependencies into a cross-modal vision-text KG with dynamic visual attributes.
At timestep $t$, we embed the results of the visual perception process as attributes to the corresponding entity nodes in $\mathcal{G}_\text{init}$, which include the spatial and conditional information ($V_\text{env}(t)$ and $V_\text{inv}(t)$) of the observed environment. This approach cross-modally integrates the static text-modal KG with dynamic spatial and conditional visual context, enhancing the agent’s understanding of the environment by linking real-time visual perception to the semantic entities represented in the KG. The resulted cross-modal vision-text graph is denoted as $\mathcal{G} = (\mathcal{V}, \mathcal{E})$.

\subsection{Graph-based Information Retrieval}
\label{PEKR}

We retrieve task-related cross-modal information from the cross-modal KG to support the reasoning and decision-making processes of the agent and reduce the computational resources occupied by redundant information. However, the lightweight design of the KG results in insufficient descriptive text, weakening traditional semantic similarity-based retrieval. Therefore, we propose a retrieval-based graph-pooling strategy for accessing the information stored within the introduced KG.

\noindent\textbf{Prompt synthesis.} For a task $s$, we define a task description $T_\text{td}(s)$ to guide the agent's focus on task-relevant information, such as ``In Minecraft, I am focusing on chopping logs and need to make reasonable action choices based on the position of logs on the screen and my crosshair position.'' To support reasoning, we include CoT $T_\text{cot}(s)$~\cite{CoT}, like ``Is the crosshair close enough to the target?'', ``Does the crosshair need further alignment?'', or ``Do you need to move closer to the target?''. The synthetic prompt for task $s$ can be defined as $\mathcal{I}(s)=\{T_\text{td}(s), T_\text{cot}(s)\}$, which guides the information retrieval process and prompts the agent’s next action.
\begin{figure}[t]
  \centering
  \setlength{\abovecaptionskip}{5pt}
  \includegraphics[width=\linewidth]{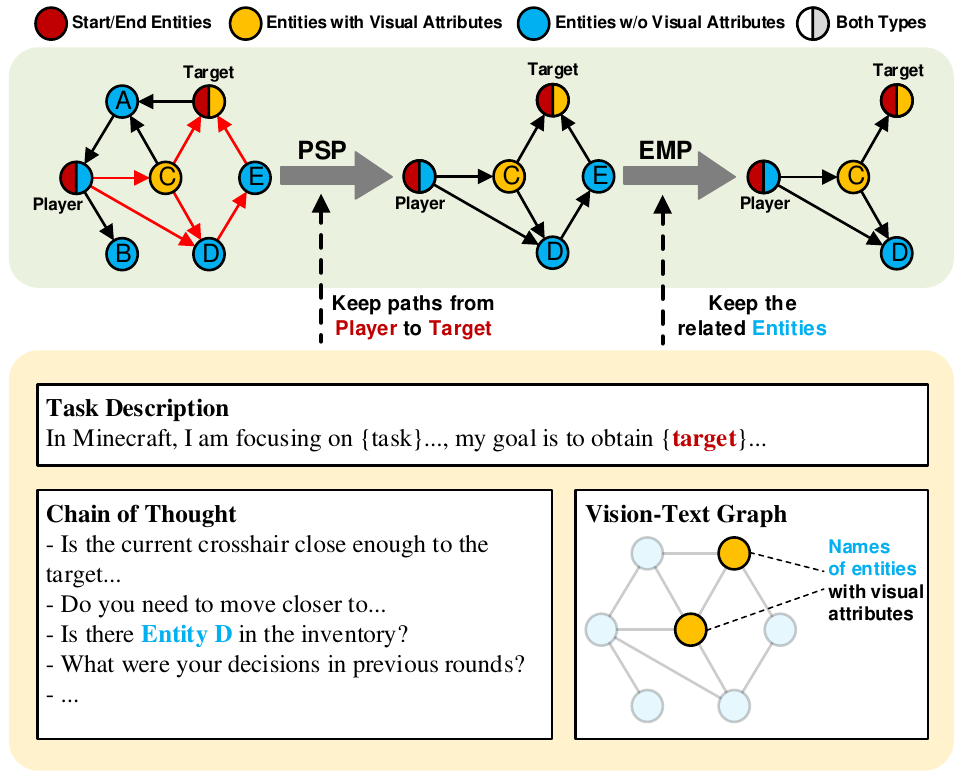}
  \caption{\textbf{Retrieval-based pooling.} It first employs path searching pooling (PSP) to retain paths from the ``Player'' node to the task-specific ``Target'' node in the KG. Subsequently, entity matching pooling (EMP) preserves entities referenced in the task prompt and those with visual attributes in the dynamic vision-text graph.  The pooled graph is textualized and input to the LLM, providing the agent with factual dependencies and real-time visual information.}
  \label{KG-Pool}
  \vspace{-0.1in}
\end{figure}

\noindent\textbf{Retrieval-based pooling.} Due to the limited textual attributes in the constructed lightweight graph $\mathcal{G}=(\mathcal{V},\mathcal{E})$, traditional information retrieval methods relying on semantic similarity ~\cite{ren2025sca3d,ren2025enhanced} face robustness issues~\cite{GraphRAG}. To address this, we propose a retrieval-based pooling mechanism for the graph, as shown in Figure~\ref{KG-Pool}, which includes path searching pooling (PSP) and entity matching pooling (EMP). Let $p$ represent a path between two nodes in $\mathcal{V}$. The PSP retains $p \in \mathcal{P}_\text{P-T}(s)$, where $\mathcal{P}_\text{P-T}(s)$ denotes the set of paths from the player node $v_\text{player}$ to the task target node $v_\text{target}(s)$, capturing global dependencies for task $s$. Each path $p \in \mathcal{P}_\text{P-T}(s)$ consists of a node set $\mathcal{V}(p)$ and an edge set $\mathcal{E}(p)$. The nodes $\mathcal{V}_\text{global}(s)$ and edges $\mathcal{E}_\text{global}(s)$ along all related paths for task $s$ can be expressed as:
\vspace{-5pt} 
\begin{equation}
\mathcal{V}_\text{global}(s) = \bigcup_{p \in \mathcal{P}_{\text{P-T}}(s)} \mathcal{V}(p),
\end{equation}
\begin{equation}
\mathcal{E}_\text{global}(s) = \bigcup_{p \in \mathcal{P}_{\text{P-T}}(s)} \mathcal{E}(p).
\end{equation}
\noindent Following PSP, the EMP checks whether each node in $\mathcal{V}_\text{global}(s)$ appears in the synthetic prompt $\mathcal{I}(s)$ and visual attributes at timestep $t$, and retains the corresponding nodes $\mathcal{V}_\text{local}(s,t) \in \mathcal{V}_\text{global}(s)$ and their edges $\mathcal{E}_\text{local}(s,t) \in \mathcal{E}_\text{global}(s)$ to retrieve dynamic local dependencies that adapt to the real-time task progress. After pooling, the KG $\mathcal{G'}(s,t) = (\mathcal{V}_\text{local}(s,t), \mathcal{E}_\text{local}(s,t))$ contains information that most relevant to the agent's state at timestep $t$, which is then textualized as input to the LLM.

\subsection{Desktop-level Skill Library}

Previous works that employ LLMs as policies typically utilize APIs (e.g., MineFlayer) to construct skill libraries for agents, with each agent interacting with the environment in high-level action spaces to accomplish complex tasks~\cite{Voyager}. However, it limits the agent's ability to generalize, as not all environments provide APIs~\cite{GROOT}. A potential improvement is enabling interaction through MNK operations, yet existing works that adopt this method rely on action libraries tailored to specific simulators, restricting their transferability~\cite{JARVIS-1,GROOT,ROCKET-1}.

To address this limitation, we develop a desktop-level skill library $\mathcal{L}$ based on the concept of general computer control~\cite{Cradle}, utilizing PyAutoGUI in hybrid action spaces~\cite{JARVIS-1}. This library encompasses low-level MNK operations (e.g., pressing a key or dragging the mouse) and their combinations (e.g., crafting items). By enabling the agent to act directly within the Minecraft desktop client through MNK inputs, the library removes the dependency on simulators or environment-specific APIs. Examples of skill functions within the library are shown on the right side of Figure~\ref{pipe}. The agent can access the library, retrieve the necessary skill functions, and autonomously generate input parameters based on the observable visual information.

\subsection{Memory Stack} 

Given the recency sensitivity of decision-making (i.e., recent decisions matter more) and the causality of continuous decisions in virtual games, we design a memory module based on the Last-In-First-Out (LIFO) stack storage concept in physical computer systems, termed the ``memory stack''. The LIFO property prioritizes the recent memory and ensures the agent can query the decision history continuously. At each timestep, the agent pushes its latest decision to the top of the memory stack. At timestep $t$, a query $q(t)$ with specified recall steps prompts the stack to pop decision histories $\mathcal{H}(t) = M(q(t))$ from top to bottom, allowing the agent to recall decisions from the most recent to earlier ones with controllable recall steps.

\subsection{Agent Workflow Modeling}
Let $\pi_\theta$ denote a parameterized LLM. At timestep $t$ for task $s$, the agent uses $\pi_\theta$ as policy to predict an executable action $\mathcal{A}(t)$ (a skill function and its parameters) based on the synthetic prompt $\mathcal{P}(s,t)$, the pooled knowledge graph $\mathcal{G'}(s,t)$, decision histories $\mathcal{H}(t)$, and the accessible skill library $\mathcal{L}$. This workflow can be modeled as:
\vspace{-5pt} 
\begin{equation}
\mathcal{A}(t)=\pi_\theta(\mathcal{I}(s), \mathcal{G'}(s,t), \mathcal{H}(t), \mathcal{L}).
\end{equation}

\vspace{-5pt} 
\noindent After $\mathcal{A}_t$ is executed in the Minecraft
environment $Env$, the observable space at the next timestep can be expressed as:
\vspace{-5pt} 
\begin{equation}
\mathcal{O}(t+1)=Env(\mathcal{A}(t)).
\end{equation}

\begin{table*}[t]
  \caption{\textbf{Comparison with other non-API-based baselines.} The comparison includes the scale of the training dataset and the GPU VRAM threshold required for integrating domain knowledge, and the success rate of milestone goals for ``obtain diamond''. `*' indicates that base models are finetuned on domain-specific text, but the dataset and GPU resources scale were not disclosed. `-' denotes not available publicly.}
  \centering
  \vspace{-5pt} 
  \resizebox{0.95\linewidth}{!}{
  \begin{tabular}{l|c|c|c|ccccccccc}
    \toprule
     Method  & Venue & Dataset Scale& GPU VRAM  &\adjustbox{valign=m}{\includegraphics[width=0.5cm]{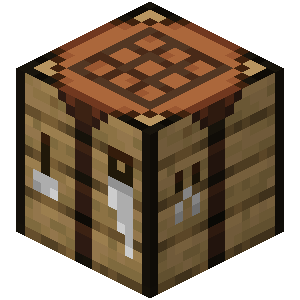}} & \adjustbox{valign=m}{\includegraphics[width=0.5cm]{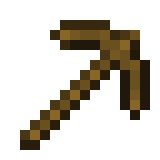}}& \adjustbox{valign=m}{\includegraphics[width=0.5cm]{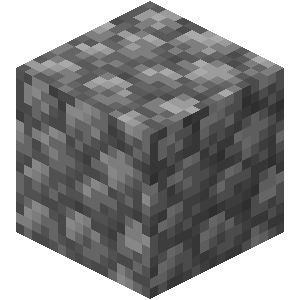}} & \adjustbox{valign=m}{\includegraphics[width=0.5cm]{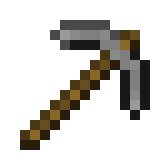}}& \adjustbox{valign=m}{\includegraphics[width=0.5cm]{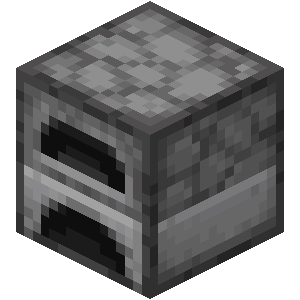}} & \adjustbox{valign=m}{\includegraphics[width=0.5cm]{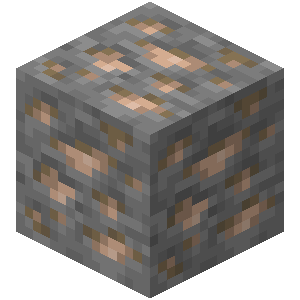}}& \adjustbox{valign=m}{\includegraphics[width=0.5cm]{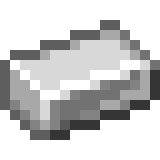}} & \adjustbox{valign=m}{\includegraphics[width=0.5cm]{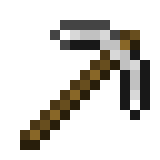}}&\adjustbox{valign=m}{\includegraphics[width=0.5cm]{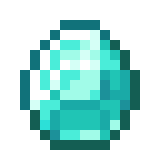}}\\
     \midrule
     STEVE-1 &NeurIPS'23 &160M frames & 192 GB &0.04&0.04&-&0.00&0.00&0.00&0.00&0.00&0.00\\
     GROOT &ICLR'24 &1.6B frames& 384 GB &  0.05&0.05&-&0.00&0.00&0.00&0.00&0.00&0.00\\
     DEPS &NeurIPS'23 &*& * & 0.90&0.80&-&0.80&0.70&-&-&0.10&0.01\\
     Omni JARVIS &NeurIPS'24 &990M tokens& 640 GB &  0.95&0.95&0.82&-&-&0.82&0.32&0.32&0.08\\
     JARVIS-1 &TPAMI'24 &* & * &  0.97&0.96&0.95&0.94&0.94&0.57&0.55&0.34&0.09\\
     VPT &NeurIPS'22&140B frames& 23040 GB &  0.99&0.99&0.99&0.98&0.78&~\textbf{0.84}&0.60&0.57&0.15\\
    ROCKET-1 &CVPR'25 &1.6B frames& * &  \textbf{1.00}&\textbf{1.00}&\textbf{1.00}&-&\textbf{1.00}&-&-&-&0.25\\
    \midrule
     \rowcolor{skyblue!40}\textbf{VistaWise (Ours)} & \textbf{EMNLP'25} &\textbf{471 frames}& \textbf{24 GB}
 & \textbf{1.00}&\textbf{1.00}&\textbf{1.00}&\textbf{1.00}&\textbf{1.00}&0.73&\textbf{0.73}&\textbf{0.73}&\textbf{0.33}\\
    \bottomrule
  \end{tabular}
  }
\label{GeT}
\vspace{-0.15in}
\end{table*}

\section{Experiment}

\subsection{Dataset and Implementation}
\noindent\textbf{Dataset.} We collect a small-scale dataset\footnote{The dataset is available at \url{https://drive.google.com/file/d/1QXGtSJJWw4emKB8RLGYUvq_fNEDCxhP_/view?usp=sharing}} to train an object detection model for Minecraft, defining 23 types of visual entities. From Minecraft gameplay videos, we extract 471 frames and annotate 3,304 instances using X-AnyLabeling. This dataset is used to finetune a pretrained object detection model for application in Minecraft environments. Additional details are provided in Appendix \ref{app:dataset}.

\noindent\textbf{Implementation details.} We use the pretrained YOLOv10-L~\cite{YOLOv10} object detection model for entity perception, finetuned with Minecraft data on an L4 GPU with 24G VRAM. The dataset is split into training and testing sets in a 9:1 ratio, with finetuning conducted over 150 epochs, a batch size of 16, and an input image size of 768. Other settings follow YOLO's default configurations in the Ultralytics package. The empirical thresholds in Sec.\ref{VPPS} are $k_w=110$ and $k_h=275$. We use GPT-4o~\cite{GPT-4o} as the LLM policy. The agent acts in the Vanilla 1.11.2 Minecraft client on a Windows 10 machine, equipped with a 1080p resolution monitor, a 3060 Ti GPU with 8G VRAM, an i7-12700F CPU, and 32GB of RAM. Unless otherwise specified, experimental results are based on 15 repeated trials. The setup of the game environment and details of the skill library are provided in Appendix \ref{app:Game Environment} and \ref{app:Skill Library}.

\subsection{Comparison against other methods}
We compare our method with 7 non-API-based baselines\footnote{We exclude comparisons with works using environmental APIs (e.g., MineFlayer) for perception and action.}. 
These methods includes: STEVE-1~\cite{STEVE-1}, GROOT~\cite{GROOT}, DEPS~\cite{DEPS}, Omni JARVIS~\cite{OmniJARVIS}, JARVIS-1~\cite{JARVIS-1}, VPT~\cite{VPT}, and ROCKET-1~\cite{ROCKET-1}.
Results for STEVE-1 and GROOT are taken from Omni JARVIS, while other results are sourced from the respective original papers.

\noindent\textbf{Performance on ``obtain diamond''.} We test our method with the ultimate goal of ``obtaining diamond,'' a classic challenge for agents in Minecraft. Table~\ref{GeT} compares our performance with others across 9 sequential sub-goals leading to this objective: obtain crafting table \adjustbox{valign=m}{\includegraphics[width=0.4cm]{1.png}} $\rightarrow$ obtain wooden pickaxe \adjustbox{valign=m}{\includegraphics[width=0.4cm]{2.png}} $\rightarrow$ ... $\rightarrow$ obtain diamond \adjustbox{valign=m}{\includegraphics[width=0.4cm]{9.png}}. These sub-goals must be executed sequentially, with failure at any stage resulting in the failure of subsequent goals. The results demonstrate that our method achieves state-of-the-art (SOTA) performance on the ultimate goal and most of the sub-goals.

\noindent\textbf{Efficiency.} As shown in Table~\ref{GeT}, we finetune the object detection model using only 471 frames extracted from game videos and 24 GB GPU VRAM, whereas other methods typically rely on at least 160M frames with 192 GB VRAM or 990M tokens of text with 64 GB VRAM. Despite the substantially smaller dataset scale and VRAM usage, we achieve SOTA performance. It demonstrates the effectiveness of our approach in reducing development costs related to data collection and training.

\begin{table}[t]
  \caption{\textbf{Ablation study on textual entity attributes} in the KG on the success rate of milestone goals.}
  \centering
  \vspace{-5pt} 
  \resizebox{0.9\linewidth}{!}{
  \begin{tabular}{l|cccccc}
    \toprule
     Attributes& \adjustbox{valign=m}{\includegraphics[width=0.5cm]{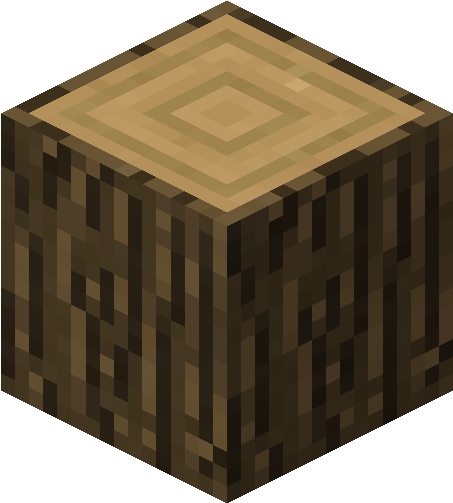}}& \adjustbox{valign=m}{\includegraphics[width=0.5cm]{2.png}}& \adjustbox{valign=m}{\includegraphics[width=0.5cm]{4.png}} & \adjustbox{valign=m}{\includegraphics[width=0.5cm]{5.png}}& \adjustbox{valign=m}{\includegraphics[width=0.5cm]{8.png}}& \adjustbox{valign=m}{\includegraphics[width=0.5cm]{9.png}}\\
     \midrule
     Full information&1.00& 1.00&1.00 &0.80&0.80 &0.27 \\
     Names only&1.00&1.00 &1.00&0.73&0.73&0.33 \\
    \bottomrule
\end{tabular}
}
\label{AEA}
\vspace{-0.15in}
\end{table}

\begin{figure}[t]
  \centering
    \vspace{-8pt} 
  \resizebox{1\linewidth}{!}{
  \subfloat{\hspace{-2.8mm} 
  \includegraphics[width=0.5\linewidth]{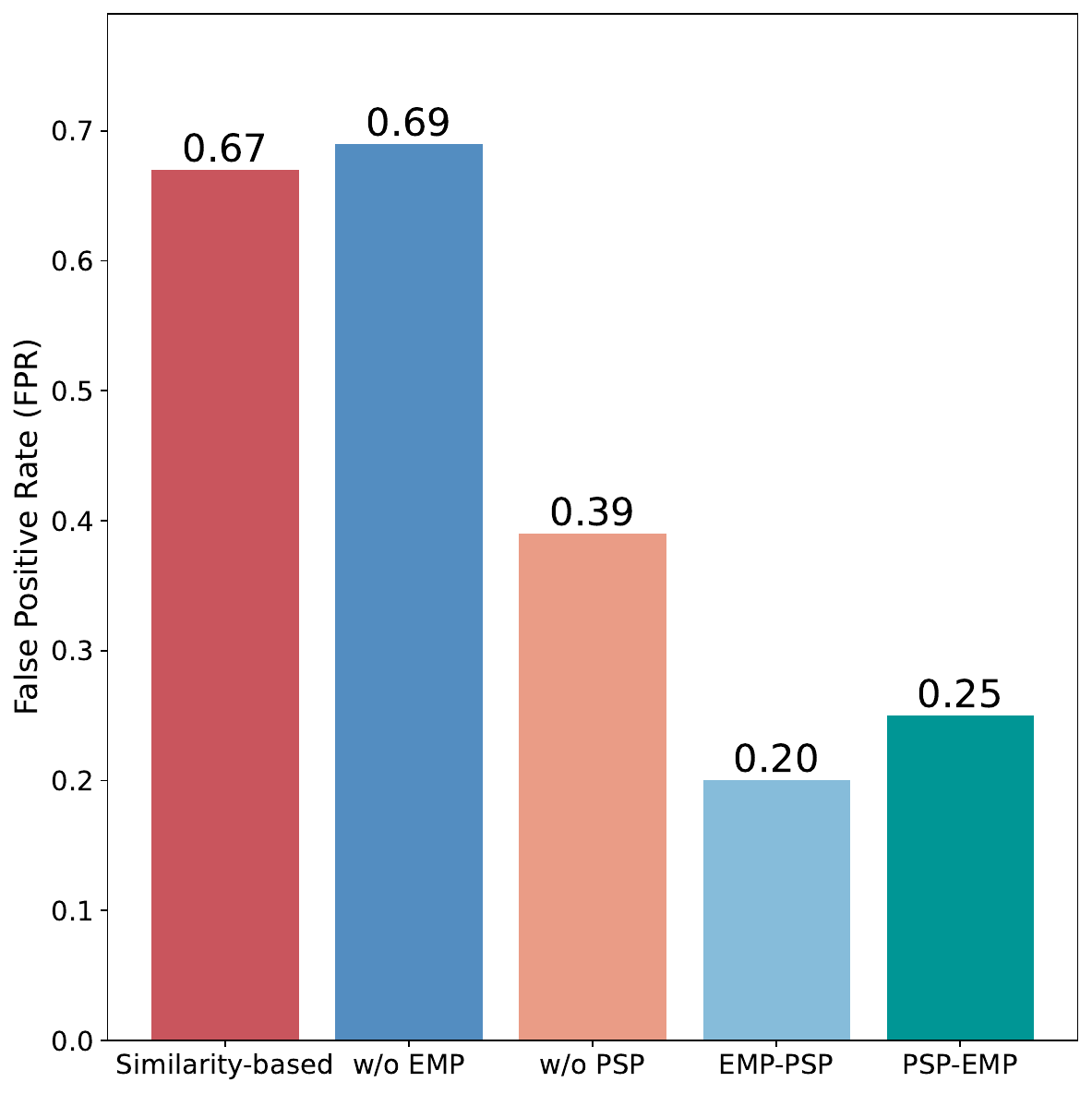}}
  \hfill
  \subfloat{\includegraphics[width=0.5\linewidth]{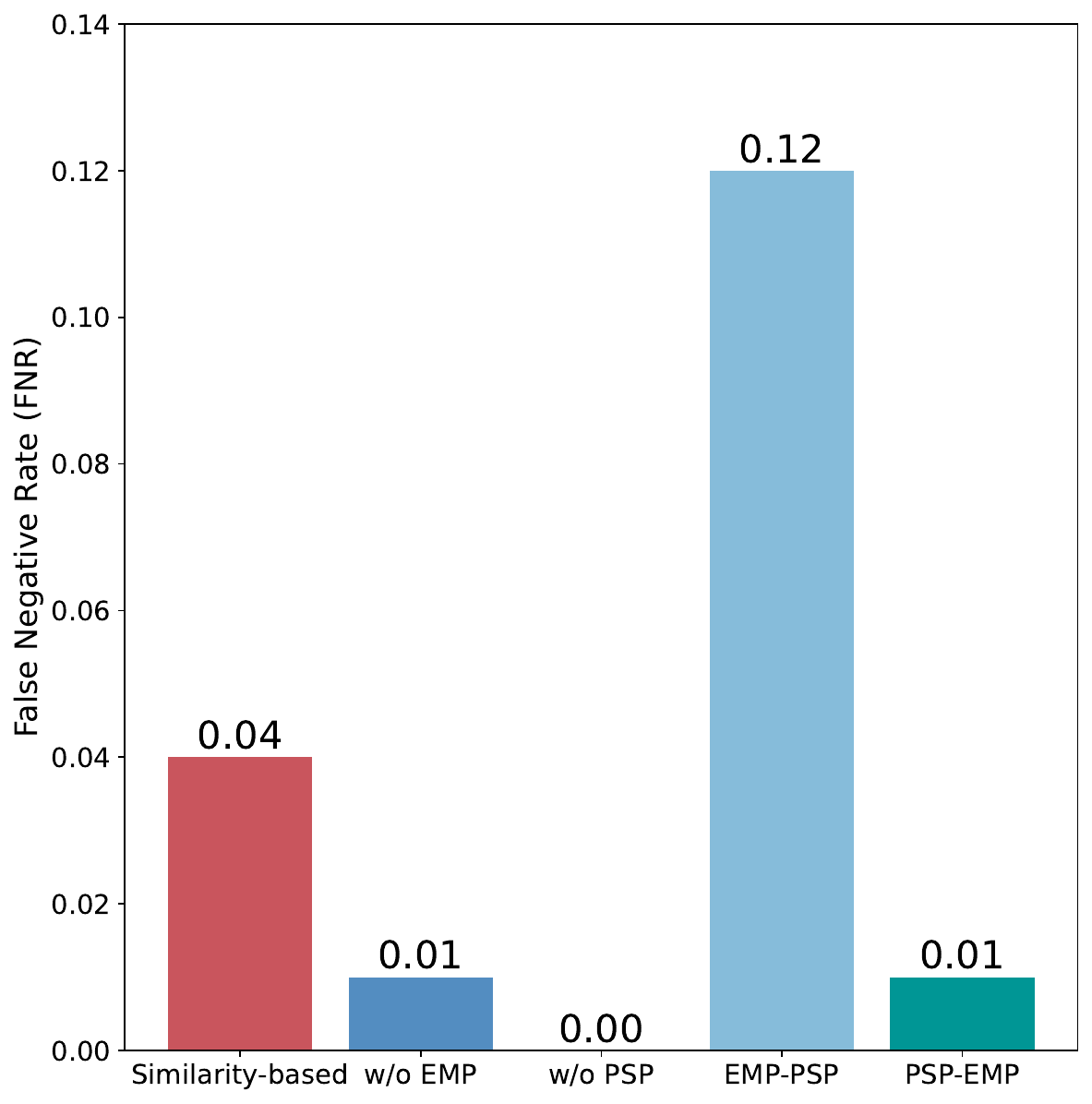}}
}
  \caption{\textbf{Ablation study on information retrieval strategies.} \textbf{(Left)} False Positive Rate (FPR), the proportion of redundant information in the retrieved results. \textbf{(Right)} False Negative Rate (FNR), the proportion of missed information to all that should have been retrieved. Lower FPR and FNR indicate better retrieval. ``Similarity-based'', ``EMP'', and ``PSP'' refer to the similarity-based strategy, entity matching pooling, and path searching pooling, respectively, while ``EMP-PSP'' and ``PSP-EMP'' denote their execution order.}
  \label{KRS}
  \vspace{-0.15in}
\end{figure}

\subsection{Ablation Study}
\textbf{Entity attributes.} In Sec.\ref{PEKR}, we construct a KG to integrate external textual knowledge for the agent, reducing hallucinations in complex crafting tasks. To minimize token overhead, we retain only entity names and exclude other information (e.g., background knowledge). As shown in Table \ref{AEA}, the agent successfully understands entity dependencies despite the omission of full information, with no crafting failures. Variations in success rates between the two ablation groups stem from the random distribution of in-game resources.

\noindent\textbf{Knowledge retrieval strategies.} Figure~\ref{KRS} compares similarity-based and pooling-based retrieval strategies in our lightweight KG using False Positive Rate (FPR) and False Negative Rate (FNR). The similarity-based strategy employs a BERT model (all-MiniLM-L6-v2) to embed prompts and graph attributes, and the cosine similarity between them to retrieve relevant entities. We also examine the roles of path pooling, entity pooling, and their order. The results indicate that the pooling strategy outperforms similarity-based retrieval in the introduced KG. Moreover, performing entity pooling first disrupts the node-edge structure, potentially losing critical paths. Thus, considering both FPR and FNR, the optimal strategy is path pooling followed by entity pooling.

\noindent\textbf{LLM policies and visual perception.} Table~\ref{LLMbackbone} evaluates several common MLLMs~\cite{Gemini,Qwen,GPT-4o} within our framework, assessing agent performance with and without object detection models or MLLM visual capabilities. Results show that without accurate visual entity information from object detection, relying solely on MLLM's visual capability is inadequate for complex tasks. Besides, as object detection provides sufficient information, additional visual input does not significantly affect performance. It indicates that using an object detection model can effectively replace MLLM's visual capability and enhance the agent's spatial perception.

\noindent\textbf{Memory stack, CoT and KG.} We examine the effectiveness of several components in Table~\ref{abla}. The results demonstrate that the memory stack is crucial for handling complex tasks, CoT serves as the foundation for decision-making, and KG contributes to enhancing stability in crafting tasks.

\begin{table}[t]
  \caption{\textbf{Ablation study on LLM policies and visual perception methods.} `V' denotes leveraging the visual capabilities of the MLLM for image understanding, while `OD' refers to using an object detection model. ``$a/b$'' indicates that the agent successfully achieved the goal $a$ times out of $b$ consecutive attempts.}
  \centering
  \vspace{-4pt} 
  \resizebox{0.95\linewidth}{!}{
  \begin{tabular}{l|c|ccccc}
    \toprule
     LLM Policy  & Method& \adjustbox{valign=m}{\includegraphics[width=0.5cm]{0.png}} & \adjustbox{valign=m}{\includegraphics[width=0.5cm]{2.png}}& \adjustbox{valign=m}{\includegraphics[width=0.5cm]{4.png}} & \adjustbox{valign=m}{\includegraphics[width=0.5cm]{8.png}}& \adjustbox{valign=m}{\includegraphics[width=0.5cm]{9.png}}\\
     \midrule
     &V&2/3&0/3&0/3&0/3&0/3\\
     \rowcolor{skyblue!15} \cellcolor{white}Gemini-1.5-pro&OD&\textbf{3/3}&\textbf{3/3}&\textbf{3/3}&\textbf{2/3}&\textbf{1/3}\\
     \rowcolor{skyblue!15} \cellcolor{white}
     &V+OD&\textbf{3/3}&\textbf{3/3}&\textbf{3/3}&\textbf{2/3}&\textbf{1/3}\\
     \midrule
     &V&0/3&0/3&0/3&0/3&0/3\\
     \rowcolor{skyblue!15} \cellcolor{white}Qwen-VL-Max&OD&\textbf{3/3}&0/3&0/3&0/3&0/3\\
     \rowcolor{skyblue!15} \cellcolor{white}
     &V+OD&\textbf{3/3}&0/3&0/3&0/3&0/3\\
     \midrule
     &V&\textbf{3/3}&0/3&0/3&0/3&0/3\\
     \rowcolor{skyblue!15} \cellcolor{white}GPT-4o&OD&\textbf{3/3}&\textbf{3/3}&\textbf{3/3}&\textbf{2/3}&\textbf{1/3}\\
     \rowcolor{skyblue!15} \cellcolor{white}
     &V+OD&\textbf{3/3}&\textbf{3/3}&\textbf{3/3}&\textbf{2/3}&\textbf{1/3}\\
    \bottomrule
  \end{tabular}
  }
\label{LLMbackbone}
\vspace{-0.05in}
\end{table}

\begin{table}[t]
  \caption{\textbf{Ablation study on memory stack, CoT and KG.} ``$a/b$'' indicates that the agent successfully achieved the goal $a$ times out of $b$ consecutive attempts.}
  \centering
    \vspace{-5pt} 
  \resizebox{0.9\linewidth}{!}{
  \begin{tabular}{l|ccccc}
    \toprule
     Method& \adjustbox{valign=m}{\includegraphics[width=0.5cm]{0.png}}& \adjustbox{valign=m}{\includegraphics[width=0.5cm]{2.png}}& \adjustbox{valign=m}{\includegraphics[width=0.5cm]{4.png}} & \adjustbox{valign=m}{\includegraphics[width=0.5cm]{8.png}}& \adjustbox{valign=m}{\includegraphics[width=0.5cm]{9.png}}\\
     \midrule
     w/o Memory Stack&\textbf{3/3}&\textbf{3/3} &\textbf{3/3} &0/3 &0/3 \\
     w/o CoT&2/3&0/3 &0/3 &0/3 &0/3 \\
     w/o KG&\textbf{3/3}&2/3&2/3&1/3 &0/3 \\
     \midrule
     \rowcolor{skyblue!40} Full&\textbf{3/3}&\textbf{3/3} &\textbf{3/3} &\textbf{2/3} &\textbf{1/3} \\
    \bottomrule
\end{tabular}
}
\label{abla}
\vspace{-0.15in}
\end{table}

\begin{figure}[t]
  \centering
  \includegraphics[width=0.95\linewidth]{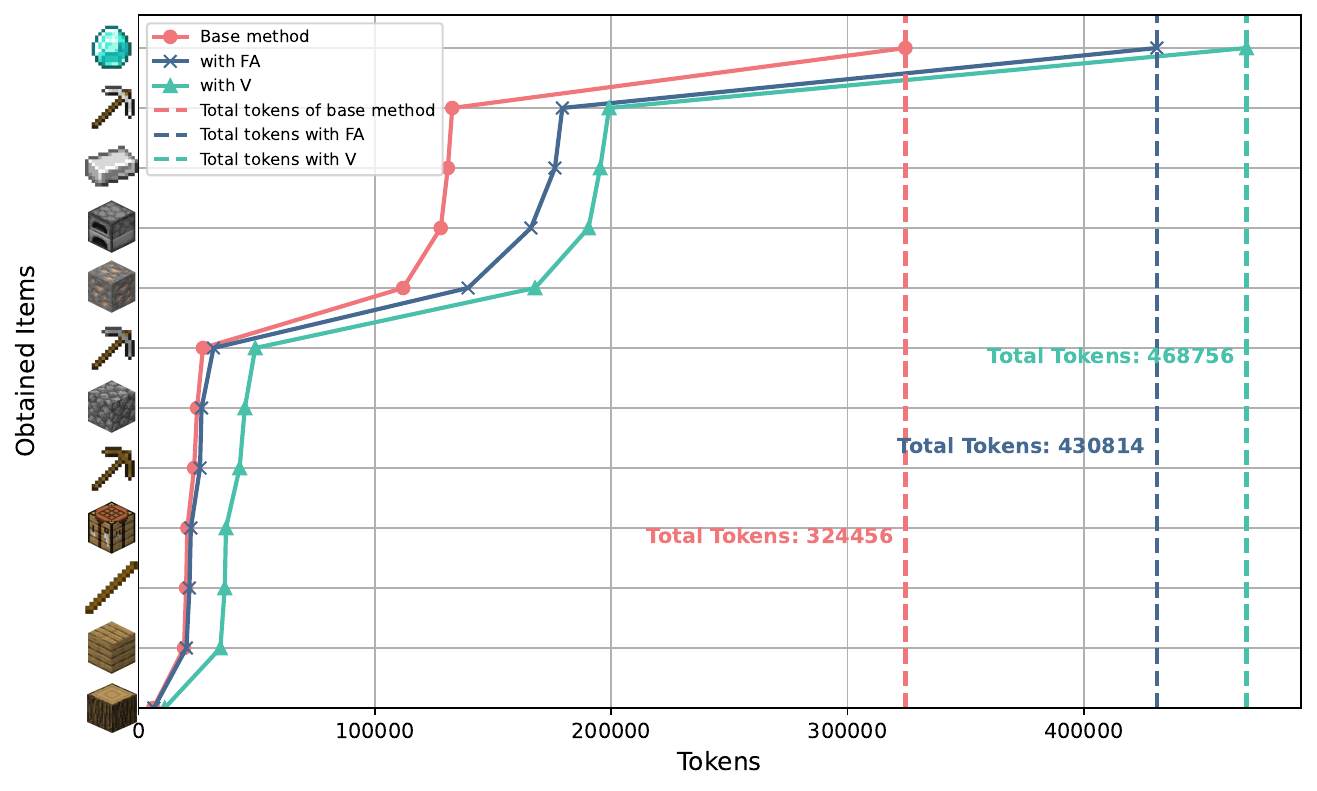}
    \vspace{-8pt} 
  \caption{\textbf{The tokens consumed by our proposed agent to successfully achieve various goals.} ``VistaWise'' is our standard framework proposed in Sec.\ref{method}, while ``with FA'' and ``with V'' indicate the addition of full graph attributes (Table~\ref{AEA}) and the use of visual input (Table~\ref{LLMbackbone}) to the standard framework, respectively.}
  \label{TokensC}
    \vspace{-0.15in}
\end{figure}

\subsection{Inference Costs}
\label{sec::Inference Costs}
In addition to the costs associated with model training, managing inference overhead is crucial to enhancing the cost-efficiency of LLM-based agents. As illustrated in Figure~\ref{TokensC}, we quantify this overhead through the token consumption required to achieve various goals. For the ultimate goal of ``obtain diamond,'' omitting full entity attributes during external knowledge injection and disabling the MLLM's visual capabilities leads to a reduction in token consumption by 30.6\% and 41.0\% respectively, while the agent performance shows no significant degradation as indicated in Tables~\ref{AEA} and~\ref{LLMbackbone}.
Moreover, by leveraging the lightweight graph-based representations of cross-modal information, our framework achieves budget-friendly inference. Based on GPT-4o pricing for 160 iterations, our framework incurs a cost of $\sim\$1.28$. In comparison, the earlier cost-transparent\footnote{https://github.com/MineDojo/Voyager/blob/main/FAQ.md} LLM policy-based agent Voyager, under the same conditions, costs around $\sim\$25$, achieving a 94.9\% reduction in cost. These results validate the cost-effective nature of our method.

\section{Conclusion}

We present VistaWise, a cost-effective agent framework for open-world embodied decision-making in Minecraft. With cross-modal domain-specific knowledge through retrieving external textual dependencies and training a dedicated object detection model, VistaWise minimizes the need for extensive domain-specific training data. It constructs a cross-modal KG combining visual information and textual dependencies for better task understanding. It also includes a retrieval-based pooling strategy to extract task-relevant data from the KG, and a desktop-level skill library to control the Minecraft desktop client directly via MNK inputs. VistaWise effectively reduces development costs while maintaining high performance, offering a novel and efficient solution for virtual open-world agents.

\section*{Limitations}

Utilizing an LLM as the policy within VistaWise impacts the agent's real-time performance, as each action iteration necessitates waiting for LLM inference and, if applicable, a server response. While pausing the virtual environment’s processes synchronizes the agent's decision-making with the environment's timescale, it results in increased time for the agent to complete tasks.

\section*{Acknowledgments}
This research is supported by the National Natural Science Foundation of China (No. 62406267), Tencent Rhino-Bird Focused Research Program and the Guangzhou Municipal Science and Technology Project (No. 2025A04J4070).

\bibliography{custom}

@article{Odyssey,
  title={Odyssey: Empowering Minecraft Agents with Open-World Skills},
  author={Liu, Shunyu and Li, Yaoru and Zhang, Kongcheng and Cui, Zhenyu and Fang, Wenkai and Zheng, Yuxuan and Zheng, Tongya and Song, Mingli},
  journal={arXiv preprint arXiv:2407.15325},
  year={2024}
}

@article{generalist,
  title={A Generalist Agent},
  author={Reed, Scott and Zolna, Konrad and Parisotto, Emilio and Colmenarejo, Sergio G{\'o}mez and Novikov, Alexander and Barth-maron, Gabriel and Gim{\'e}nez, Mai and Sulsky, Yury and Kay, Jackie and Springenberg, Jost Tobias and others},
  journal={Transactions on Machine Learning Research},
  year={2022}
}

@article{VPT,
  title={Video pretraining (vpt): Learning to act by watching unlabeled online videos},
  author={Baker, Bowen and Akkaya, Ilge and Zhokov, Peter and Huizinga, Joost and Tang, Jie and Ecoffet, Adrien and Houghton, Brandon and Sampedro, Raul and Clune, Jeff},
  journal={Advances in Neural Information Processing Systems},
  volume={35},
  pages={24639--24654},
  year={2022}
}

@article{STEVE-1,
  title={Steve-1: A generative model for text-to-behavior in minecraft},
  author={Lifshitz, Shalev and Paster, Keiran and Chan, Harris and Ba, Jimmy and McIlraith, Sheila},
  journal={Advances in Neural Information Processing Systems},
  volume={36},
  pages={69900--69929},
  year={2023}
}

@inproceedings{MineRL,
  title={MineRL: a large-scale dataset of minecraft demonstrations},
  author={Guss, William H and Houghton, Brandon and Topin, Nicholay and Wang, Phillip and Codel, Cayden and Veloso, Manuela and Salakhutdinov, Ruslan},
  booktitle={Proceedings of the 28th International Joint Conference on Artificial Intelligence},
  pages={2442--2448},
  year={2019}
}

@article{MineDOJO,
  title={Minedojo: Building open-ended embodied agents with internet-scale knowledge},
  author={Fan, Linxi and Wang, Guanzhi and Jiang, Yunfan and Mandlekar, Ajay and Yang, Yuncong and Zhu, Haoyi and Tang, Andrew and Huang, De-An and Zhu, Yuke and Anandkumar, Anima},
  journal={Advances in Neural Information Processing Systems},
  volume={35},
  pages={18343--18362},
  year={2022}
}

@inproceedings{enhancedP,
  title={Enhanced poet: Open-ended reinforcement learning through unbounded invention of learning challenges and their solutions},
  author={Wang, Rui and Lehman, Joel and Rawal, Aditya and Zhi, Jiale and Li, Yulun and Clune, Jeffrey and Stanley, Kenneth},
  booktitle={International conference on machine learning},
  pages={9940--9951},
  year={2020},
  organization={PMLR}
}

@article{proto,
  title={Proto: Program-guided transformer for program-guided tasks},
  author={Zhao, Zelin and Samel, Karan and Chen, Binghong and others},
  journal={Advances in neural information processing systems},
  volume={34},
  pages={17021--17036},
  year={2021}
}

@inproceedings{STEVE,
  title={See and think: Embodied agent in virtual environment},
  author={Zhao, Zhonghan and Chai, Wenhao and Wang, Xuan and Li, Boyi and Hao, Shengyu and Cao, Shidong and Ye, Tian and Wang, Gaoang},
  booktitle={European Conference on Computer Vision},
  pages={187--204},
  year={2025},
  organization={Springer}
}

@article{LARM,
  title={LARM: Large Auto-Regressive Model for Long-Horizon Embodied Intelligence},
  author={Li, Zhuoling and Xu, Xiaogang and Xu, Zhenhua and Lim, SerNam and Zhao, Hengshuang},
  journal={arXiv preprint arXiv:2405.17424},
  year={2024}
}

@inproceedings{LLaMA-Rider,
  title={LLaMA-Rider: Spurring Large Language Models to Explore the Open World},
  author={Feng, Yicheng and Wang, Yuxuan and Liu, Jiazheng and Zheng, Sipeng and Lu, Zongqing},
  booktitle={Findings of the Association for Computational Linguistics: NAACL 2024},
  pages={4705--4724},
  year={2024}
}

@inproceedings{MP5,
  title={Mp5: A multi-modal open-ended embodied system in minecraft via active perception},
  author={Qin, Yiran and Zhou, Enshen and Liu, Qichang and Yin, Zhenfei and Sheng, Lu and Zhang, Ruimao and Qiao, Yu and Shao, Jing},
  booktitle={2024 IEEE/CVF Conference on Computer Vision and Pattern Recognition (CVPR)},
  pages={16307--16316},
  year={2024},
  organization={IEEE}
}

@article{
Voyager,
title={Voyager: An Open-Ended Embodied Agent with Large Language Models},
author={Guanzhi Wang and Yuqi Xie and Yunfan Jiang and Ajay Mandlekar and Chaowei Xiao and Yuke Zhu and Linxi Fan and Anima Anandkumar},
journal={Transactions on Machine Learning Research},
issn={2835-8856},
year={2024},
url={https://openreview.net/forum?id=ehfRiF0R3a},
note={}
}

@article{GITM,
  title={Ghost in the minecraft: Generally capable agents for open-world environments via large language models with text-based knowledge and memory},
  author={Zhu, Xizhou and Chen, Yuntao and Tian, Hao and Tao, Chenxin and Su, Weijie and Yang, Chenyu and Huang, Gao and Li, Bin and Lu, Lewei and Wang, Xiaogang and others},
  journal={arXiv preprint arXiv:2305.17144},
  year={2023}
}

@inproceedings{DEPS,
  title={Describe, explain, plan and select: interactive planning with large language models enables open-world multi-task agents},
  author={Wang, Zihao and Cai, Shaofei and Chen, Guanzhou and Liu, Anji and Ma, Xiaojian and Liang, Yitao and CraftJarvis, Team},
  booktitle={Proceedings of the 37th International Conference on Neural Information Processing Systems},
  pages={34153--34189},
  year={2023}
}

@inproceedings{GROOT,
  title={GROOT: Learning to Follow Instructions by Watching Gameplay Videos},
  author={Cai, Shaofei and Zhang, Bowei and Wang, Zihao and Ma, Xiaojian and Liu, Anji and Liang, Yitao},
  booktitle={The Twelfth International Conference on Learning Representations},
  year={2024}
}

@article{JARVIS-1,
  title={Jarvis-1: Open-world multi-task agents with memory-augmented multimodal language models},
  author={Wang, Zihao and Cai, Shaofei and Liu, Anji and Jin, Yonggang and Hou, Jinbing and Zhang, Bowei and Lin, Haowei and He, Zhaofeng and Zheng, Zilong and Yang, Yaodong and others},
  journal={IEEE Transactions on Pattern Analysis and Machine Intelligence},
  year={2024},
  publisher={IEEE}
}

@article{ROCKET-1,
  title={ROCKET-1: Master Open-World Interaction with Visual-Temporal Context Prompting},
  author={Cai, Shaofei and Wang, Zihao and Lian, Kewei and Mu, Zhancun and Ma, Xiaojian and Liu, Anji and Liang, Yitao},
  journal={arXiv preprint arXiv:2410.17856},
  year={2024}
}

@inproceedings{Cradle,
  title={Cradle: Empowering Foundation Agents towards General Computer Control},
  author={Tan, Weihao and Zhang, Wentao and Xu, Xinrun and Xia, Haochong and Ding, Gang and Li, Boyu and Zhou, Bohan and Yue, Junpeng and Jiang, Jiechuan and Li, Yewen and others},
  booktitle={NeurIPS 2024 Workshop on Open-World Agents},
  year={2024}
}

@article{CoT,
  title={Chain-of-thought prompting elicits reasoning in large language models},
  author={Wei, Jason and Wang, Xuezhi and Schuurmans, Dale and Bosma, Maarten and Xia, Fei and Chi, Ed and Le, Quoc V and Zhou, Denny and others},
  journal={Advances in neural information processing systems},
  volume={35},
  pages={24824--24837},
  year={2022}
}

@article{Qwen,
  title={Qwen-vl: A versatile vision-language model for understanding, localization, text reading, and beyond},
  author={Bai, Jinze and Bai, Shuai and Yang, Shusheng and Wang, Shijie and Tan, Sinan and Wang, Peng and Lin, Junyang and Zhou, Chang and Zhou, Jingren},
  journal={arXiv preprint arXiv:2308.12966},
  volume={1},
  number={2},
  pages={3},
  year={2023}
}

@article{Gemini,
  title={Gemini: a family of highly capable multimodal models},
  author={Team, Gemini and Anil, Rohan and Borgeaud, Sebastian and Alayrac, Jean-Baptiste and Yu, Jiahui and Soricut, Radu and Schalkwyk, Johan and Dai, Andrew M and Hauth, Anja and Millican, Katie and others},
  journal={arXiv preprint arXiv:2312.11805},
  year={2023}
}

@article{GPT-4o,
  title={Gpt-4 technical report},
  author={Achiam, Josh and Adler, Steven and Agarwal, Sandhini and Ahmad, Lama and Akkaya, Ilge and Aleman, Florencia Leoni and Almeida, Diogo and Altenschmidt, Janko and Altman, Sam and Anadkat, Shyamal and others},
  journal={arXiv preprint arXiv:2303.08774},
  year={2023}
}

@article{YOLOv10,
  title={Yolov10: Real-time end-to-end object detection},
  author={Wang, Ao and Chen, Hui and Liu, Lihao and Chen, Kai and Lin, Zijia and Han, Jungong and others},
  journal={Advances in Neural Information Processing Systems},
  volume={37},
  pages={107984--108011},
  year={2024}
}

@article{OmniJARVIS,
  title={Omnijarvis: Unified vision-language-action tokenization enables open-world instruction following agents},
  author={Wang, Zihao and Cai, Shaofei and Mu, Zhancun and Lin, Haowei and Zhang, Ceyao and Liu, Xuejie and Li, Qing and Liu, Anji and Ma, Xiaojian Shawn and Liang, Yitao},
  journal={Advances in Neural Information Processing Systems},
  volume={37},
  pages={73278--73308},
  year={2024}
}

@article{GraphRAG,
  title={From local to global: A graph rag approach to query-focused summarization},
  author={Edge, Darren and Trinh, Ha and Cheng, Newman and Bradley, Joshua and Chao, Alex and Mody, Apurva and Truitt, Steven and Larson, Jonathan},
  journal={arXiv preprint arXiv:2404.16130},
  year={2024}
}

@article{RAG,
  title={Retrieval-augmented generation for knowledge-intensive nlp tasks},
  author={Lewis, Patrick and Perez, Ethan and Piktus, Aleksandra and Petroni, Fabio and Karpukhin, Vladimir and Goyal, Naman and K{\"u}ttler, Heinrich and Lewis, Mike and Yih, Wen-tau and Rockt{\"a}schel, Tim and others},
  journal={Advances in Neural Information Processing Systems},
  volume={33},
  pages={9459--9474},
  year={2020}
}

@inproceedings{RAG2,
  title={Active Retrieval Augmented Generation},
  author={Jiang, Zhengbao and Xu, Frank F and Gao, Luyu and Sun, Zhiqing and Liu, Qian and Dwivedi-Yu, Jane and Yang, Yiming and Callan, Jamie and Neubig, Graham},
  booktitle={Proceedings of the 2023 Conference on Empirical Methods in Natural Language Processing},
  pages={7969--7992},
  year={2023}
}

@article{KGRAG,
  title={Knowledge Graph Retrieval-Augmented Generation for LLM-based Recommendation},
  author={Wang, Shijie and Fan, Wenqi and Feng, Yue and Ma, Xinyu and Wang, Shuaiqiang and Yin, Dawei},
  journal={arXiv preprint arXiv:2501.02226},
  year={2025}
}

@article{GRAG,
  title={GRAG: Graph Retrieval-Augmented Generation},
  author={Hu, Yuntong and Lei, Zhihan and Zhang, Zheng and Pan, Bo and Ling, Chen and Zhao, Liang},
  journal={arXiv preprint arXiv:2405.16506},
  year={2024}
}

@article{GraphRAG2,
  title={Retrieval-Augmented Generation with Graphs (GraphRAG)},
  author={Han, Haoyu and Wang, Yu and Shomer, Harry and Guo, Kai and Ding, Jiayuan and Lei, Yongjia and Halappanavar, Mahantesh and Rossi, Ryan A and Mukherjee, Subhabrata and Tang, Xianfeng and others},
  journal={arXiv preprint arXiv:2501.00309},
  year={2024}
}

@inproceedings{SAM,
  title={Segment anything},
  author={Kirillov, Alexander and Mintun, Eric and Ravi, Nikhila and Mao, Hanzi and Rolland, Chloe and Gustafson, Laura and Xiao, Tete and Whitehead, Spencer and Berg, Alexander C and Lo, Wan-Yen and others},
  booktitle={Proceedings of the IEEE/CVF international conference on computer vision},
  pages={4015--4026},
  year={2023}
}

@inproceedings{
wang2023selfconsistency,
title={Self-Consistency Improves Chain of Thought Reasoning in Language Models},
author={Xuezhi Wang and Jason Wei and Dale Schuurmans and Quoc V Le and Ed H. Chi and Sharan Narang and Aakanksha Chowdhery and Denny Zhou},
booktitle={The Eleventh International Conference on Learning Representations },
year={2023},
url={https://openreview.net/forum?id=1PL1NIMMrw}
}

@article{pan2024unifying,
  title={Unifying large language models and knowledge graphs: A roadmap},
  author={Pan, Shirui and Luo, Linhao and Wang, Yufei and Chen, Chen and Wang, Jiapu and Wu, Xindong},
  journal={IEEE Transactions on Knowledge and Data Engineering},
  volume={36},
  number={7},
  pages={3580--3599},
  year={2024},
  publisher={IEEE}
}

@inproceedings{
jiang2023unikgqa,
title={Uni{KGQA}: Unified Retrieval and Reasoning for Solving Multi-hop Question Answering Over Knowledge Graph},
author={Jinhao Jiang and Kun Zhou and Xin Zhao and Ji-Rong Wen},
booktitle={The Eleventh International Conference on Learning Representations },
year={2023},
url={https://openreview.net/forum?id=Z63RvyAZ2Vh}
}

@inproceedings{jiang2023reasoninglm,
  title={ReasoningLM: Enabling Structural Subgraph Reasoning in Pre-trained Language Models for Question Answering over Knowledge Graph},
  author={Jiang, Jinhao and Zhou, Kun and Zhao, Wayne Xin and Li, Yaliang and Wen, Ji-Rong},
  booktitle={Proceedings of the 2023 Conference on Empirical Methods in Natural Language Processing},
  pages={3721--3735},
  year={2023}
}

@inproceedings{zhang2022subgraph,
  title={Subgraph Retrieval Enhanced Model for Multi-hop Knowledge Base Question Answering},
  author={Zhang, Jing and Zhang, Xiaokang and Yu, Jifan and Tang, Jian and Tang, Jie and Li, Cuiping and Chen, Hong},
  booktitle={Proceedings of the 60th Annual Meeting of the Association for Computational Linguistics (Volume 1: Long Papers)},
  pages={5773--5784},
  year={2022}
}

@inproceedings{
jiang2023structgpt,
title={Struct{GPT}: A General Framework for Large Language Model to Reason over Structured Data},
author={Jinhao Jiang and Kun Zhou and zican Dong and KeMing Ye and Xin Zhao and Ji-Rong Wen},
booktitle={The 2023 Conference on Empirical Methods in Natural Language Processing},
year={2023},
url={https://openreview.net/forum?id=R635gF7lXD}
}

@inproceedings{
sun2024thinkongraph,
title={Think-on-Graph: Deep and Responsible Reasoning of Large Language Model on Knowledge Graph},
author={Jiashuo Sun and Chengjin Xu and Lumingyuan Tang and Saizhuo Wang and Chen Lin and Yeyun Gong and Lionel Ni and Heung-Yeung Shum and Jian Guo},
booktitle={The Twelfth International Conference on Learning Representations},
year={2024},
url={https://openreview.net/forum?id=nnVO1PvbTv}
}

@article{jiang2024kg,
  title={Kg-agent: An efficient autonomous agent framework for complex reasoning over knowledge graph},
  author={Jiang, Jinhao and Zhou, Kun and Zhao, Wayne Xin and Song, Yang and Zhu, Chen and Zhu, Hengshu and Wen, Ji-Rong},
  journal={arXiv preprint arXiv:2402.11163},
  year={2024}
}

@inproceedings{
wang2025reasoning,
title={Reasoning of Large Language Models over Knowledge Graphs with Super-Relations},
author={Song Wang and Junhong Lin and Xiaojie Guo and Julian Shun and Jundong Li and Yada Zhu},
booktitle={The Thirteenth International Conference on Learning Representations},
year={2025},
url={https://openreview.net/forum?id=rTCJ29pkuA}
}

@inproceedings{baek2023knowledge,
  title={Knowledge-Augmented Language Model Prompting for Zero-Shot Knowledge Graph Question Answering},
  author={Baek, Jinheon and Aji, Alham and Saffari, Amir},
  booktitle={The 61st Annual Meeting Of The Association For Computational Linguistics},
  year={2023}
}

@article{ren2025sca3d,
  title={SCA3D: Enhancing Cross-modal 3D Retrieval via 3D Shape and Caption Paired Data Augmentation},
  author={Ren, Junlong and Wu, Hao and Xiong, Hui and Wang, Hao},
  journal={arXiv preprint arXiv:2502.19128},
  year={2025}
}

@misc{mei2025survey,
      title={A Survey of Context Engineering for Large Language Models}, 
      author={Lingrui Mei and Jiayu Yao and Yuyao Ge and Yiwei Wang and Baolong Bi and Yujun Cai and Jiazhi Liu and Mingyu Li and Zhong-Zhi Li and Duzhen Zhang and Chenlin Zhou and Jiayi Mao and Tianze Xia and Jiafeng Guo and Shenghua Liu},
      year={2025},
      eprint={2507.13334},
      archivePrefix={arXiv},
      primaryClass={cs.CL},
      url={https://arxiv.org/abs/2507.13334}, 
}

@article{mei2024slang,
  title={SLANG: New Concept Comprehension of Large Language Models},
  author={Mei, Lingrui and Liu, Shenghua and Wang, Yiwei and Bi, Baolong and Cheng, Xueqi},
  journal={EMNLP 2024},
  year={2024}
}

@article{fu2023sgcn,
  title={SGCN: a multi-order neighborhood feature fusion landform classification method based on superpixel and graph convolutional network},
  author={Fu, Honghao and Shen, Yilang and Liu, Yuxuan and Li, Jingzhong and Zhang, Xiang},
  journal={International Journal of Applied Earth Observation and Geoinformation},
  volume={122},
  pages={103441},
  year={2023},
  publisher={Elsevier}
}

@article{mei2024hiddenguard,
  title={HiddenGuard: Fine-Grained Safe Generation with Specialized Representation Router},
  author={Mei, Lingrui and Liu, Shenghua and Wang, Yiwei and Bi, Baolong and Yuan, Ruibin and Cheng, Xueqi},
  journal={arXiv preprint arXiv:2410.02684},
  year={2024}
}

@article{fu2024dp,
  title={DP-IQA: Utilizing Diffusion Prior for Blind Image Quality Assessment in the Wild},
  author={Fu, Honghao and Wang, Yufei and Yang, Wenhan and Kot, Alex C and Wen, Bihan},
  journal={arXiv preprint arXiv:2405.19996},
  year={2024}
}

@inproceedings{fu2025brainvis,
  title={BrainVis: Exploring the bridge between brain and visual signals via image reconstruction},
  author={Fu, Honghao and Wang, Hao and Chin, Jing Jih and Shen, Zhiqi},
  booktitle={ICASSP 2025-2025 IEEE International Conference on Acoustics, Speech and Signal Processing (ICASSP)},
  pages={1--5},
  year={2025},
  organization={IEEE}
}

@article{liu2025step1x,
  title={Step1x-edit: A practical framework for general image editing},
  author={Liu, Shiyu and Han, Yucheng and Xing, Peng and Yin, Fukun and Wang, Rui and Cheng, Wei and Liao, Jiaqi and Wang, Yingming and Fu, Honghao and Han, Chunrui and others},
  journal={arXiv preprint arXiv:2504.17761},
  year={2025}
}

@article{mei2024not,
  title={"Not Aligned" is Not" Malicious": Being Careful about Hallucinations of Large Language Models' Jailbreak},
  author={Mei, Lingrui and Liu, Shenghua and Wang, Yiwei and Bi, Baolong and Mao, Jiayi and Cheng, Xueqi},
  journal={COLING 2025},
  year={2024}
}

@article{wang2025text,
  title={Text Speaks Louder than Vision: ASCII Art Reveals Textual Biases in Vision-Language Models},
  author={Wang, Zhaochen and Hooi, Bryan and Wang, Yiwei and Yang, Ming-Hsuan and Huang, Zi and Cai, Yujun},
  journal={arXiv preprint arXiv:2504.01589},
  year={2025}
}

@inproceedings{wang2021mixup,
  title={Mixup for node and graph classification},
  author={Wang, Yiwei and Wang, Wei and Liang, Yuxuan and Cai, Yujun and Hooi, Bryan},
  booktitle={Proceedings of the Web Conference 2021},
  pages={3663--3674},
  year={2021}
}

@article{wang2025cure,
  title={Cure or Poison? Embedding Instructions Visually Alters Hallucination in Vision-Language Models},
  author={Wang, Zhaochen and Wang, Yiwei and Cai, Yujun},
  journal={arXiv preprint arXiv:2508.01678},
  year={2025}
}

@article{gu2025hdtcnet,
  title={HDTCNet: A hybrid-dimensional convolutional network for multivariate time series classification},
  author={Gu, Yongli and Yan, Xiang and Qin, Hanlin and Akhtar, Naveed and Yuan, Shuai and Fu, Honghao and Yang, Shuowen and Mian, Ajmal},
  journal={Pattern Recognition},
  pages={111837},
  year={2025},
  publisher={Elsevier}
}

@inproceedings{li2024vulnerability,
  title={Vulnerability of LLMs to Vertically Aligned Text Manipulations},
  author={Li, Zhecheng and Wang, Yiwei and Hooi, Bryan and Cai, Yujun and Xiong, Zhen and Peng, Nanyun and Chang, Kai-Wei},
  booktitle={Association for Computational Linguistics ACL, 2025.},
  year={2024}
}

@inproceedings{li2024drs,
  title={DRS: Deep Question Reformulation With Structured Output},
  author={Li, Zhecheng and Wang, Yiwei and Hooi, Bryan and Cai, Yujun and Peng, Nanyun and Chang, Kai-Wei},
  booktitle={Association for Computational Linguistics ACL, 2025.},
  year={2024}
}

@inproceedings{li2025texture,
  title={Texture or Semantics? Vision-Language Models Get Lost in Font Recognition},
  author={Li, Zhecheng and Song, Guoxian and Cai, Yujun and Xiong, Zhen and Yuan, Junsong and Wang, Yiwei},
  booktitle={Conference on Language Modeling COLM, 2025.},
  year={2025}
}

@article{xiong2025mapping,
  title={Mapping the Minds of LLMs: A Graph-Based Analysis of Reasoning LLM},
  author={Xiong, Zhen and Cai, Yujun and Li, Zhecheng and Wang, Yiwei},
  journal={arXiv preprint arXiv:2505.13890},
  year={2025}
}

@article{ren2025wamo,
  title={WaMo: Wavelet-Enhanced Multi-Frequency Trajectory Analysis for Fine-Grained Text-Motion Retrieval},
  author={Ren, Junlong and Zhang, Gangjian and Fu, Honghao and Wu, Pengcheng and Wang, Hao},
  journal={arXiv preprint arXiv:2508.03343},
  year={2025}
}

@inproceedings{ren2025diversified,
  title={Diversified Augmentation with Domain Adaptation for Debiased Video Temporal Grounding},
  author={Ren, Junlong and Zhang, Gangjian and Sun, Haifeng and Wang, Hao},
  booktitle={ICASSP 2025-2025 IEEE International Conference on Acoustics, Speech and Signal Processing (ICASSP)},
  pages={1--5},
  year={2025},
  organization={IEEE}
}

@article{ren2025enhanced,
  title={Enhanced Cross-modal 3D Retrieval via Tri-modal Reconstruction},
  author={Ren, Junlong and Wang, Hao},
  journal={arXiv preprint arXiv:2504.01476},
  year={2025}
}

@article{ren2025exploiting,
  title={Exploiting Inter-Sample Correlation and Intra-Sample Redundancy for Partially Relevant Video Retrieval},
  author={Ren, Junlong and Zhang, Gangjian and Hu, Yu and Shu, Jian and Wang, Hao},
  journal={arXiv preprint arXiv:2504.19637},
  year={2025}
}

@INPROCEEDINGS{10888525,
  author={Zhang, Zheng and Lan, Yihuai and Chen, Yangsen and Wang, Lei and Wang, Xiang and Wang, Hao},
  booktitle={ICASSP 2025 - 2025 IEEE International Conference on Acoustics, Speech and Signal Processing (ICASSP)}, 
  title={DVM: Towards Controllable LLM Agents in Social Deduction Games}, 
  year={2025},
  volume={},
  number={},
  pages={1-5},
  doi={10.1109/ICASSP49660.2025.10888525}}

@misc{zhang2025multimindenhancingwerewolfagents,
      title={MultiMind: Enhancing Werewolf Agents with Multimodal Reasoning and Theory of Mind}, 
      author={Zheng Zhang and Nuoqian Xiao and Qi Chai and Deheng Ye and Hao Wang},
      year={2025},
      eprint={2504.18039},
      archivePrefix={arXiv},
      primaryClass={cs.AI},
      url={https://arxiv.org/abs/2504.18039}, 
}

@article{zhang2024defending,
  title={Defending multimodal backdoored models by repulsive visual prompt tuning},
  author={Zhang, Zhifang and He, Shuo and Wang, Haobo and Shen, Bingquan and Feng, Lei},
  journal={NeurIPS},
  year={2025}
}

@inproceedings{zhang2025tuning,
  title={Tuning vision-language models with candidate labels by prompt alignment},
  author={Zhang, Zhifang and Niu, Yuwei and Liu, Xin and Li, Beibei},
  booktitle={DASFAA},
  year={2025},
}

@inproceedings{zhang2026test,
  title={Test-Time Attention Purification for Backdoored Large Vision Language Models},
  author={Zhang, Zhifang and Yang, Bojun and He, Shuo and Chen, Weitong and Zhang, Wei Emma and Maennel, Olaf and Feng, Lei and and Xu, Miao},
  booktitle={CVPR},
  year={2026},
}

@inproceedings{ti2025towards,
  title={Towards Reverse Engineering of Language Models: A Survey},
  author={Ti, Xinpeng and Ye, Wentao and Zhang, Zhifang and Zhao, Junbo and Yao, Chang and Feng, Lei and Wang, Haobo},
  booktitle={Findings of the Association for Computational Linguistics: EMNLP 2025},
  year={2025}
}

@article{zhang2025tokenswap,
  title={Tokenswap: Backdoor attack on the compositional understanding of large vision-language models},
  author={Zhang, Zhifang and Tao, Qiqi and Lv, Jiaqi and Zhao, Na and Feng, Lei and Zhou, Joey Tianyi},
  journal={arXiv preprint arXiv:2509.24566},
  year={2025}
}

@article{zhang2025improving,
  title={Improving generalizability and undetectability for targeted adversarial attacks on multimodal pre-trained models},
  author={Zhang, Zhifang and Zhang, Jiahan and Zhou, Shengjie and Wei, Qi and He, Shuo and Liu, Feng and Feng, Lei},
  journal={arXiv preprint arXiv:2509.19994},
  year={2025}
}

@misc{liu2025correlationcausationmaxpoolingbasedmultiinstance,
      title={From Correlation to Causation: Max-Pooling-Based Multi-Instance Learning Leads to More Robust Whole Slide Image Classification}, 
      author={Xin Liu and Weijia Zhang and Wei Tang and Thuc Duy Le and Jiuyong Li and Lin Liu and Min-Ling Zhang},
      year={2025},
      eprint={2408.09449},
      archivePrefix={arXiv},
      primaryClass={cs.CV},
      url={https://arxiv.org/abs/2408.09449}, 
}

@inproceedings{liu2025hacsurv,
  title={HACSurv: A Hierarchical Copula-Based Approach for Survival Analysis with Dependent Competing Risks},
  author={Liu, Xin and Zhang, Weijia and Zhang, Min-Ling},
  booktitle={International Conference on Artificial Intelligence and Statistics},
  pages={3079--3087},
  year={2025},
  organization={PMLR}
}

@article{gu2023orsi,
  title={Orsi salient object detection via bidimensional attention and full-stage semantic guidance},
  author={Gu, Yubin and Xu, Honghui and Quan, Yueqian and Chen, Wanjun and Zheng, Jianwei},
  journal={IEEE Transactions on Geoscience and Remote Sensing},
  volume={61},
  pages={1--13},
  year={2023},
  publisher={IEEE}
}

@article{gu2025optical,
  title={Optical remote sensing image salient object detection via bidirectional cross-attention and attention restoration},
  author={Gu, Yubin and Chen, Siting and Sun, Xiaoshuai and Ji, Jiayi and Zhou, Yiyi and Ji, Rongrong},
  journal={Pattern Recognition},
  volume={164},
  pages={111478},
  year={2025},
  publisher={Elsevier}
}

@inproceedings{gu2025acl,
  title={Acl: Activating capability of linear attention for image restoration},
  author={Gu, Yubin and Meng, Yuan and Ji, Jiayi and Sun, Xiaoshuai},
  booktitle={Proceedings of the Computer Vision and Pattern Recognition Conference},
  pages={17913--17923},
  year={2025}
}

@article{gu2025sfir,
  title={Sfir: Optimizing spatial and frequency domains for image restoration},
  author={Gu, Yubin and Meng, Yuan and Chen, Siting and Ji, Jiayi and Sun, Xiaoshuai and Ruan, Weijian and Ji, Rongrong},
  journal={Pattern Recognition},
  pages={112188},
  year={2025},
  publisher={Elsevier}
}

@article{meng2025wavelet,
  title={Wavelet-based learning and optimized sampling for image deraining},
  author={Meng, Yuan and Gu, Yubin and Sun, Xiaoshuai and Ji, Jiayi and Ruan, Weijian and Ji, Rongrong},
  journal={Pattern Recognition},
  pages={112782},
  year={2025},
  publisher={Elsevier}
}

\appendix

\newpage
\section{The Explanation of Icons}
\label{app:icons}

We explain the goals represented by the icons mentioned in the main paper in Table~\ref{icon}.

\begin{table}[h]
  \caption{The explanation of icons mentioned in the main paper.}
  \centering
  \resizebox{1\linewidth}{!}{
  \begin{tabular}{c|p{6cm}}
    \toprule
    Icons (Goals) & Explanations\\
    \midrule
    \raisebox{-0.5\height}{\adjustbox{valign=m}{\includegraphics[width=0.5cm]{0.png}}}&Obtain log. Chop a tree using the hands or an axe to collect wood logs.\\
    \midrule
    \raisebox{-0.5\height}{\adjustbox{valign=m}{\includegraphics[width=0.5cm]{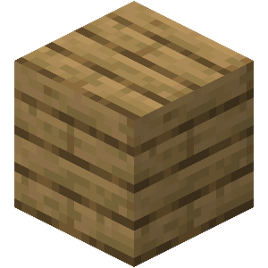}}}&Obtain wooden plank. Craft wooden planks from logs.\\
    \midrule
    \raisebox{-0.5\height}{\adjustbox{valign=m}{\includegraphics[width=0.5cm]{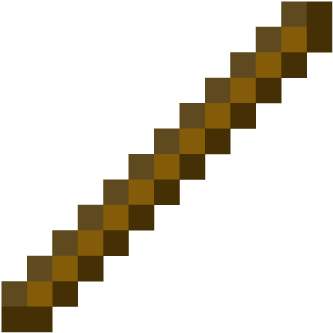}}}&Obtain stick. Craft sticks from wooden planks.\\
    \midrule
    \raisebox{-0.5\height}{\adjustbox{valign=m}{\includegraphics[width=0.5cm]{1.png}}}&Obtain crafting table. Craft a crafting table from wooden planks.\\
    \midrule
    \raisebox{-1\height}{\adjustbox{valign=m}{\includegraphics[width=0.5cm]{2.png}}}&Obtain wooden pickaxe. Use wooden planks and sticks to craft a wooden pickaxe at the crafting table.\\
    \midrule
    \raisebox{-1\height}{\adjustbox{valign=m}{\includegraphics[width=0.5cm]{3.png}}}&Obtain cobblestone. Mine stone blocks with wooden pickaxe to collect cobblestone.\\
    \midrule
    \raisebox{-1\height}{\adjustbox{valign=m}{\includegraphics[width=0.5cm]{4.png}}}&Obtain stone pickaxe. Craft a stone pickaxe using cobblestones and sticks at the crafting table.\\
    \midrule
    \raisebox{-0.5\height}{\adjustbox{valign=m}{\includegraphics[width=0.5cm]{5.png}}}&Obtain iron ore. Mine iron ore blocks (found underground) with stone pickaxe.\\
    \midrule
    \raisebox{-0.5\height}{\adjustbox{valign=m}{\includegraphics[width=0.5cm]{6.png}}}&Obtain furnace. Craft a furnace using cobblestones at the crafting table.\\
    \midrule
    \raisebox{-0.5\height}{\adjustbox{valign=m}{\includegraphics[width=0.5cm]{7.png}}}&Obtain iron ingot. Smelt the iron ore in a furnace using fuel.\\
    \midrule
    \raisebox{-1\height}{\adjustbox{valign=m}{\includegraphics[width=0.5cm]{8.png}}}&Obtain iron pickaxe. Craft an iron pickaxe using iron ingots and sticks at the crafting table.\\
    \midrule
    \raisebox{-1.5\height}{\adjustbox{valign=m}{\includegraphics[width=0.5cm]{9.png}}}&Obtain diamond, the rarest mineral in our environment. Mine diamond ore blocks (found deep underground) with an iron pickaxe.\\
    \bottomrule
\end{tabular}
}
\label{icon}
\end{table}

\section{Dataset}
\label{app:dataset}

\begin{table}[t]
  \caption{The defined visual entity types in the dataset and their annotation counts in the 471 frames of the game screen.}
  \centering
  \begin{tabular}{l|c}
    \toprule
    Entities & Annotations\\
    \midrule
    bedrock&59\\
    coal\_icon&110\\
    coal\_ore&101\\
    cobblestone\_icon&249\\
    crafting\_table\_icon&106\\
    diamond\_icon&79\\
    diamond\_ore&50\\
    diamond\_pickaxe\_icon&15\\
    dirt\_icon&115\\
    furnace\_icon&67\\
    iron\_icon&103\\
    iron\_ore&141\\
    iron\_ore\_icon&76\\
    iron\_pickaxe\_icon&38\\
    lava&15\\
    log\_icon&361\\
    non\_cobblestone\_icon&152\\
    plank\_icon&446\\
    stick\_icon&181\\
    stone\_pickaxe\_icon&72\\
    trunk&608\\
    water&37\\
    wood\_pickaxe\_icon&123\\
    \midrule
    Total&3304\\

    \bottomrule
\end{tabular}
\label{data}
\end{table}

We use X-AnyLabeling to annotate a total of 3,304 visual entities across 471 frames extracted from the gameplay video. The entity details and their corresponding annotation counts are provided in Table~\ref{data}. Entities such as trunk, plank\_icon, and log\_icon exhibit higher annotation counts, since each of them has different types of textures (such as oak, birch, etc.).

\section{Game Environment}
\label{app:Game Environment}

We use the Vanilla 1.11.2 Minecraft desktop client as the virtual environment for the agent, with the specific game world settings detailed in Table~\ref{opt}.

\begin{table}[t]
  \caption{Game world settings.}
  \centering
  \begin{tabular}{l|c}
    \toprule
    Game Options& Settings\\
    \midrule
    Game Mode& Survival\\
    Generate Structures&ON\\
    Bonus Chest&OFF\\
    Allow Cheats&ON\\
    Sea Level&63\\
    Caves&Yes\\
    Strongholds&Yes\\
    Villages&Yes\\
    Mineshafts&Yes\\
    Temples&Yes\\
    Ocean Monuments&Yes\\
    Woodland Mansions&Yes\\
    Ravines&Yes\\
    Dungeons&Yes\\
    Dungeon Count&7\\
    Water Lakes&Yes\\
    Water Lake Rarity&4\\
    Lava Lakes&No\\
    Lava Lake Rarity&80\\
    Lava Oceans&No\\
    Biome&Plains\\
    Biome Size&4\\
    River Size&4\\
    \bottomrule
\end{tabular}
\label{opt}
\end{table}

\begin{table*}[t]
  \caption{Examples from the desktop-level skill library.}
  \centering
  \resizebox{0.9\linewidth}{!}{
  \begin{tabular}{p{5.5cm}|p{11.2cm}}
    \toprule
    Skill Function & Explanation\\
    \midrule
    craft\_furnace(c\_x, c\_y) & Use cobblestone to craft a furnace. `c\_x' and `c\_y' are the x and y coordinates of the cobblestone in the player's inventory.\\
        \midrule
    craft\_iron\_pickaxe(i\_x, i\_y, s\_x, s\_y) & Use iron ingots and sticks to craft an iron pickaxe. `i\_x' and `i\_y' are the x and y coordinates of the iron ingots in the player's inventory, `s\_x' and `s\_y' are the x and y coordinates of the sticks.\\
        \midrule
    craft\_plank(l\_x, l\_y) & Use logs to craft planks. `l\_x' and `l\_y' are the x and y coordinates of the logs in the player's inventory.\\
        \midrule
    craft\_stick(p\_x, p\_y) & Use planks to craft sticks. `p\_x' and `p\_y' are the x and y coordinates of the planks in the player's inventory.\\
        \midrule
    craft\_stone\_pickaxe(c\_x, c\_y, s\_x, s\_y) & Use cobblestone and sticks to craft a stone pickaxe. `c\_x' and `c\_y' are the x and y coordinates of the cobblestone in the player's inventory, `s\_x' and `s\_y' are the x and y coordinates of the sticks.\\
        \midrule
    craft\_wood\_pickaxe(p\_x, p\_y, s\_x, s\_y) & Use planks and sticks to craft a wooden pickaxe. `p\_x' and `p\_y' are the x and y coordinates of the planks in the player's inventory, `s\_x' and `s\_y' are the x and y coordinates of the sticks.\\
        \midrule
    dig\_horizontal\_mine\_tunnels(k) & Select a pickaxe in the hotbar and use it to dig horizontal tunnels to explore ores. `k' is a number from 1 to 9 corresponding to the key for the pickaxe in the hotbar.\\
        \midrule
    dig\_vertical\_mine\_tunnels(k, d) & Select a pickaxe from the hotbar and use it to dig a vertical mine tunnel to reach (deeper) underground. `k' is a number from 1 to 9 corresponding to the key for the pickaxe in the hotbar. `d' is mouse press duration.\\
            \midrule
    mine\_log(d) & When log is within 5 meters and the crosshair is correctly aligned with the target, mine the log. `d' is mouse press duration.\\
        \midrule
    mine\_diamond\_ore(k, d) & When diamond ore is within 5 meters and the crosshair is correctly aligned with the target, mine the diamond ore. `k' is the key for the iron pickaxe in the hotbar, and `d' is mouse press duration.\\
        \midrule
    mine\_iron\_ore(k, d) & When iron ore is within 5 meters and the crosshair is correctly aligned with the target, mine the iron ore. `k' is the key for the stone pickaxe in the hotbar, and `d' is mouse press duration.\\
        \midrule
    move\_forward(d) & Move forward for a certain duration. `d' is the press duration of `w'.\\
        \midrule
    move\_item\_to\_hotbar(t\_x, t\_y) & Move a specific item from the inventory to the hotbar. `t\_x' and `t\_y' are the x and y coordinates of the target item in the player's inventory.\\
        \midrule
    place\_blocks\_underfoot(k, n) & Place cobblestone blocks underfoot to raise the player from lower to higher position. `k' is a number from 1 to 9 corresponding to the key for the cobblestone in the hotbar. `n' is the number of placed blocks.\\
            \midrule
    put\_functional\_block(k, d) & Select a functional block such as a crafting table or furnace from the hotbar and place it on the ground. `k' is a number from 1 to 9 corresponding to the key for the functional block in the hotbar. `d' is mouse press duration.\\
        \midrule
    smelt\_iron\_ore(i\_o\_x, i\_o\_y, p\_x, p\_y) & Use a furnace to smelt iron ore and obtain iron ingots. `i\_o\_x' and `i\_o\_y' are the x and y coordinates of the iron ore in the player's inventory, `p\_x' and `p\_y' are the x and y coordinates of the planks.\\
        \midrule
    turn(x, y) & Turn the view to the target. `x' and `y' are the horizontal and vertical pixel offset of the crosshair from the target, with positive or negative values.\\
        \midrule
    turn\_and\_move\_forward(d, x, y) & Firstly turn to the target, then move forward. `x' and `y' are the pixel offset of the crosshair from the target, corresponding to the horizontal and vertical directions, with positive or negative values. `d' is the press duration of the key `w'.\\
    \bottomrule
  \end{tabular}
  }
\label{sk}
\end{table*}

\section{Skill Library}
\label{app:Skill Library}
We present some important skill functions from the skill library and their explanations in Table~\ref{sk}.

The construction of the skill library is semi-automated. We first manually design a few simple template functions, then define the requirements and pass them to an LLM to generate the corresponding functions. Specifically, the human effort involved includes: (i) Designing a small number of template functions intended to cover all relevant keyboard and mouse operations in the game; (ii) Defining the requirements, including the name of each skill function, its input parameters, and brief descriptions of both the function and its parameters; (iii) Manually refining the generated functions when issues are present.
Using GPT-4o mini, we successfully generate all skill functions, with fewer than half requiring minor manual adjustments.

\section{Knowledge Graph}
Entity and relationship extraction is automated using a multi-stage process. We employ an LLM (GPT-4o) with web tool to construct the knowledge graph. Selected websites are provided to the LLM, which summarizes the knowledge (entities and their dependencies) from these sites. Due to context length limitations, each site is processed individually, and the outputs are merged into a rough draft document. The LLM is then prompted to refine this document, remove redundancy, and generate a Python-compatible knowledge graph using the NetworkX package.

While this approach produces detailed graphs, it may also introduce task-irrelevant relations, resulting in unnecessary complexity. To mitigate this, we restrict edge types to nine specific categories ("can use", "can mine", "is used to craft", "is used to produce", "can be put in/on", "is the fuel of", "includes", "can be used to mine", and "outputs") to ensure relevance and clarity. Then, a brief manual review is conducted to remove redundant or irrelevant nodes, further improving the quality of the graph. Human involvement is primarily limited to (i) selecting source websites, (ii) defining allowed relation types, and (iii) performing minor refinements.

\section{Prompt Template}
As an example, we report the prompt template for action prediction in Table \ref{prompt dependencies}.

\begin{table*}
\centering
\resizebox{0.98\textwidth}{!}
{
\begin{tabular}{p{0.95\textwidth}}
\toprule
\large\textbf{General Prompt Template}  \\

\hangindent=1em \hangafter=1 \qquad \small\texttt{In Minecraft, player is focusing on \{task\_description\}, requiring strategic action choices based on the environment and inventory status The crosshair position is at \{crosshair\_position\} in the screen. If applicable, the distance between the player and the \{target\} is \{greater than/close to/less than\} the interactable range.
}
\\

\hangindent=1em \hangafter=1 \qquad \small\texttt{
Here is some knowledge related to the current status that may help clarify item-behavior dependencies: \{retrieved\_knowledge\_graph\}.
}\\

\hangindent=1em \hangafter=1 \qquad \small\texttt{Current inventory status: \{retrieved\_attributes\_of\_observable\_inventory\}.}\\

\hangindent=1em \hangafter=1 \qquad \small\texttt{Current environment status: \{retrieved\_attributes\_of\_observable\_environment\}.}\\

\hangindent=1em \hangafter=1 \qquad \small\texttt{Available actions are defined as functions with the following formats and descriptions:
\{skill\_library\}.}\\

\hangindent=1em \hangafter=1 \qquad \small\texttt{Previous round(s) of action decision(s): \{retrieved\_memory\}.}\\

\hangindent=1em \hangafter=1 \qquad \small\texttt{Please address these questions to determine the next optimal action: \{chain\_of\_thought\}.
}\\

\hangindent=1em \hangafter=1 \qquad \small\texttt{Based on this reasoning, decide the best next action and calculate the required parameter values. The output format must be: ``Action: skill\_function(*params)''.}\\

\bottomrule
\end{tabular}
}
\caption{General prompt template for action prediction.}
\label{prompt dependencies}
\end{table*}

\end{document}